\documentclass[sensors,article,accept,moreauthors,pdftex,letterpaper]{Definitions/mdpi}

\firstpage{1} 
\pubvolume{21}
\issuenum{9}
\articlenumber{2984}
\pubyear{2021}
\copyrightyear{2021}
\externaleditor{Academic Editor: Alexander Wong} 
\datereceived{29 March 2021} 
\dateaccepted{20 April 2021} 
\datepublished{23 April 2021} 
\hreflink{https://doi.org/10.3390/s21092984} 

\usepackage{algorithmic}

\usepackage{colortbl}
\usepackage{listings}

\usepackage[normalem]{ulem}

\definecolor{mygreen}{rgb}{0,0.6,0}
\definecolor{mygray}{rgb}{0.5,0.5,0.5}
\definecolor{mymauve}{rgb}{0.58,0,0.82}

\Title{Quantization and Deployment of Deep Neural Networks on~Microcontrollers}

\TitleCitation{Quantization and Deployment of Deep Neural Networks on Microcontrollers}

\Author{\textls[-10]{Pierre-Emmanuel~Novac $^{1,}$*\orcidA, 
    Ghouthi~Boukli~Hacene $^{2,3}$\orcidB,
    Alain~Pegatoquet $^1$,
    Benoît~Miramond $^1$,
    Vincent~Gripon $^2$\orcidE}}
    
\AuthorNames{Pierre-Emmanuel Novac, Ghouthi Boukli Hacene, Alain Pegatoquet, Benoît Miramond and Vincent Gripon}

\AuthorCitation{Novac, P.-E.; Boukli Hacene, G.; Pegatoquet, A.; Miramond B.; Gripon V.}

\address{ 
$^{1}$ \quad {Université Côte d'Azur, CNRS, LEAT,} 
06903 Sophia Antipolis, France;
 Alain.Pegatoquet@univ-cotedazur.fr (A.P.); Benoit.Miramond@univ-cotedazur.fr (B.M.)\\
$^{2}$ \quad {IMT Atlantique}, 29200 Brest, France; Ghouthi.BoukliHacene@imt-atlantique.fr (G.B.H.);  \mbox{vincent.gripon@imt-atlantique.fr (V.G.)}\\
$^{3}$ \quad {MILA}, Montreal, QC H2S 3H1, Canada } 

\corres{Correspondence: Pierre-Emmanuel.Novac@univ-cotedazur.fr} 

\abstract{Embedding Artificial Intelligence into low-power devices is a challenge that has been partly overcome with recent advances in machine learning and hardware design.
Presently, deep neural networks can be deployed on embedded targets to perform different tasks such as speech recognition, object detection or human activity recognition. 
However, there is still room for optimization of deep neural networks in embedded devices. These optimizations mainly address power consumption, memory and real-time constraints, but also an easier deployment at the edge. 
Moreover, there is still a need to better understand what can be achieved for different use cases.
This work focuses on the quantization and deployment of deep neural networks on low-power 32-bit microcontrollers.
First, we outline quantization methods, relevant in the context of embedded execution on a microcontroller, Then we present a new framework for end-to-end deep neural network training, quantization and deployment.
This open-source framework, called MicroAI, is designed as an alternative to existing inference engines (TensorFlow Lite for Microcontrollers and STM32Cube.AI). Our framework can easily be adjusted and/or extended for specific use cases. 
Executions using single-precision 32-bit floating-point as well as fixed-point on 8- and 16-bit integers are supported. 
The proposed quantization method is evaluated with three different datasets (UCI-HAR, Spoken MNIST and GTSRB).  
Finally, a comparison study between MicroAI and both existing embedded inference engines is provided in terms of memory and power efficiency. On-device evaluation was done using ARM Cortex-M4F-based microcontrollers (Ambiq Apollo3 and~STM32L452RE).}


\definecolor{OliveGreen}{rgb}{0,0.6,0}

\begin{document}

\section{Introduction}
Deep Neural Networks (DNN) are widely used presently to solve a range of problems, including classification. DNN can classify all sorts of data such as audio, images or accelerometer samples for tasks such as speech recognition, object recognition or human activity recognition (HAR).

A well-known downside of DNN is its high energy consumption requirement. 
In particular, the training phase is usually based on a large amount of data processed by costly algorithms. 
Although the inference phase requires less processing power, it is still a costly process. Therefore, GPUs and ASICs are often used to perform such computations in the cloud~\cite{TPUGPUCPUBenchmark}.

However, cloud computing requires transmitting the collected data to a network server to process it and fetch the result, thus requiring permanent connectivity, causing privacy concerns as well as non-deterministic latency.
As an alternative, computations can be done at the edge on the device itself. By doing so, data do not need to be sent by the device to the cloud anymore. However, running DNN on resource-constrained devices such as a microcontroller used in Internet of Things (IoT) devices or wearables is a challenging task{~\cite{mcunet, CMSISNN, SAS2020}}.

These devices have only a very small amount of memory, often less than 1 MiB. They also run DNN algorithms several orders of magnitude more slowly than GPUs or even CPUs (see Appendix \ref{appendix:mcucpugpubench}). The reason is that microcontrollers generally rely on a general-purpose processing core that does not implement parallelization techniques such as thread-level parallelism or advanced vectorization. Moreover, microcontrollers typically run at a much lower frequency than GPUs (8~MHz to 80~MHz compared to 1~GHz to 2~GHz). Microcontrollers can also be coupled with tiny battery cells. In some cases, for example when data are collected in remote areas, they cannot even be recharged in the field.
Therefore, performing inference at the edge faces major issues in terms of real-time constraints, power consumption and memory footprint. To meet these constraints, the deployment of a DNN must respect an upper bound for one inference response time as well as an upper bound for the number of parameters of the network.

As a result, a DNN must be limited in width and depth to be deployable on a microcontroller. 
As has been observed, deeper and/or wider networks are often able to solve more complex tasks with better accuracy~\cite{EfficientNET}. As such, there is always a trade-off between memory footprint, response time, power consumption and accuracy of the model.
In a previous work~\cite{DSD2020}, we presented a trade-off between memory footprint, power consumption and accuracy when performing HAR on smart glasses. This work  showed that HAR is feasible in real time on a low-power Cortex-M4F-based microcontroller. However, we also concluded that there was room for improvement in the memory footprint and power consumption.

A technique that can provide a significant decrease in the memory footprint is based on network quantization. Quantization consists of reducing the number of bits used to encode each weight of the model, so that the total memory footprint is reduced by the same factor.
Quantization also enables the use of fixed-point rather than floating-point encoding. In other words, operations can be performed using integer rather than floating-point data types.
This is of interest because integer operations require considerably fewer computations on most processor cores, including microcontrollers.
Without a floating-point unit (FPU), floating-point instructions must be emulated in software, creating a large overhead, as was illustrated in~\cite{FPUSoftHardHybrid}. In that 
study, a comparison between software, hardware and custom hybrid FPU implementations was provided.

In this paper, we present an open-source~\cite{microaisoftware} framework, called MicroAI, to perform end-to-end training, quantization and deployment of deep neural networks on microcontrollers. The training phase relies on the well-known TensorFlow and PyTorch deep learning frameworks. Our objective is to provide a framework that is easy to adapt and extend, while maintaining a good compromise between accuracy, energy efficiency and memory footprint.

As a second contribution, we provide some comparative results using two different microcontrollers (STM32L452RE and Ambiq Apollo3) and three different inference engines (TensorFlow Lite for Microcontrollers, STM32Cube.AI and our own MicroAI). Results are compared in terms of memory footprint, inference time and power efficiency. 
Finally, we propose to apply 8-bit and 16-bit quantization methods on three datasets dealing with different modalities: acceleration and angular velocity from body-worn sensors for UCI-HAR, speech for Spoken MNIST and images for GTSRB. These datasets are light enough to be handled by a deep neural network running on a microcontroller, but still relevant for applications relying on embedded artificial intelligence.

Section \ref{sec:sota} presents the challenges of running deep neural networks on microcontrollers and recent advances in this field.
Section \ref{sec:realnumbers} describes the two common ways of representing real numbers on modern computers: floating point and fixed point. 
Section \ref{sec:nnquant} presents the methodology implemented in our MicroAI framework for deep neural network quantization. 
Section \ref{sec:deployment} details our MicroAI framework and compares it to existing solutions.
In Section \ref{sec:results}, some comparative results between our framework MicroAI and two popular embedded neural network frameworks (TensorFlow Lite for Microcontrollers and STM32Cube.AI) are given for two microcontroller platforms (Nucleo-L452RE-P and SparkFun Edge) in terms of inference time and power efficiency. The impact of our 8- and 16-bit quantization methods for three different datasets (UCI-HAR, Spoken MNIST and GTSRB) is also presented.
In Section \ref{sec:discussion} the results obtained are discussed.
Finally, Section \ref{sec:conclusion} concludes this work and discusses future perspectives.

\section{The State of the Art in Embedded Execution of Quantized Neural Networks}
\label{sec:sota}
A first family of optimization methods for embedded deep neural networks, based on network quantization, proposes to reduce the precision (i.e., the number of bits used to represent a value). {Numerous works propose to use low-bit quantization (i.e., 2 or 3 bits to quantize both weights and activations) such as PArameterized Clipping acTivation (PACT) combined with Statistics-Aware Weight Binning (SAWB)~\cite{choi2018bridging}, Learned Step Size Quantization (LSQ)~\cite{esser2019learned}, Bit-Pruning~\cite{nikolic2020bitpruning} or Differentiable Quantization of Deep Neural Networks~\cite{uhlich2019differentiable} (DQDNN).} In the extreme case, this amounts to binarizing the network parameters {\cite{BNN,rastegari2016xnor}}. Nevertheless, it is usually possible to find a good compromise between the precision and the performance of a given architecture.

Another family of methods focuses on low-cost architectures. This is notably the case of the well-known MobileNet~\cite{MobileNet} and Squeeze-Net~\cite{SqueezeNet} networks. Finding a good architecture for a given problem is difficult, and remains an open issue even if hardware constraints are not taken into account. Indeed, the combinatorial explosion of hyperparameters makes the exploration of an architecture very expensive.
These networks are often tailored to solve computer vision problems such as ImageNet~\cite{ImageNet} classification, and are therefore not well suited for general use cases or simpler problems.
In practice, many applications use a generic network such as ResNet~\cite{ResNet}. This is also the approach adopted in this work. The reason is that we want to easily make use of the same deep neural network on different kinds of data (time series, audio spectrum and image) and also simplify the implementation. 

In a third family of methods, some authors have explored the possibility of reducing the number of parameters in neural networks by identifying parts of the network that are not very useful for decision making. These parts can then be removed from the architecture. These pruning techniques make it possible to considerably reduce the number of parameters. Hence, in \cite{Han2}, the authors managed to remove up to $90\%$ of the parameters. However, the unstructured nature of the removed parameters makes it difficult to operate the network. Recently, several works have  proposed pruning in a structured way where a whole kernel, a filter or even a layer is pruned according to a specific criterion~\cite{yamamoto2018pcas,hacene2019attention,ramakrishnan2020differentiable,he2020learning}.

Finally, a last family of methods consists of identifying similar parts at various points in the architectures in order to factorize them. For example, in \cite{Han2015} the authors demonstrate that it is possible to reduce the number of parameters by replacing them with memory pointers. {Some works propose to improve accuracy by adding some learning steps and algorithms to get a more representative factorization~\cite{fard2020deep,cardinaux2020iteratively}.}

All these compression techniques make it possible to reduce the size of architectures considerably, but usually at the cost of a reduction in performance. In practice, it is also common to reduce by half or even two-thirds the memory used to store the parameters while maintaining a similar level of performance.

In this work we will focus on quantization-based compression.
To reduce the number of bits used to represent a value, quantization maps values from one set to another smaller set. The smaller set can have a constant step between its elements, in which case the quantization scheme is said to be uniform.
In the case of convolutional neural networks, the authors in \cite{GaussianApprox} show that the weights of convolutional layers typically follow a Gaussian distribution when weight decay is applied.
More generally, it has been shown that weights can closely fit Gaussian Mixture Models~\cite{LayersGMM}.
Therefore, choosing a non-uniform quantization scheme to better represent the values of the non-uniform distribution of weights would lead to a lower quantization error. 

A non-uniform quantization scheme was implemented in \cite{Bosch} on an FPGA device. In this work, instead of coding the value in a fixed-point format, only the nearest power of two is coded. Using this approach, it is possible to obtain a better resolution compared to a fixed-point representation for numbers near $0$. This approach also allows large values to be represented, but at the cost of a lower resolution. The quantization step is determined by minimizing the quantization error at the output of the layer, thus balancing the precision and the dynamic range. As the implementation relies on bit shifts rather than on integer multiplications, this solution has some benefits in terms of resource usage and latency for an FPGA target. Additionally, results show that there is a slight degradation of accuracy when using the proposed non-uniform quantization versus a uniform quantization. 

Using lower-precision computation for deep neural networks has been explored in \cite{LowPrec} where the authors compare the test error rates on various image datasets for single-precision floating point, half-precision floating point, 20-bit fixed point and their own dynamic fixed-point approach with 10 bits for activations and 12 bits for weights. In their work, it is worth noting that the training is also performed using lower-precision arithmetic.
Training with fixed-point arithmetic was presented in \cite{BackPropLimPrec} with \mbox{16-bit} weights and 8-bit inputs, causing an accuracy loss of a few percent in the evaluation of text-to-speech, parity bit computation, protein structure prediction and sonar signal classification problems.
In \cite{ImpSpeedCPU}, the authors showed that on an Intel® E5640 microprocessor with an x86 
architecture, using 8-bit integer instructions instead of floating-point instructions provided an execution speedup of more than 2, without a loss of accuracy, on a speech recognition problem. In this case the training was performed using single-precision floating-point arithmetic, and the evaluation was done after the quantization of the network~parameters.

These prior works were however mostly not concerned with embedded computing on microcontrollers. Running deep neural networks on microcontrollers began to be popular in the last few years thanks to the rise of the Internet of Things and the improved efficiency of deep neural networks.

In \cite{XpulpNN}, the authors emphasize that the instruction set architecture (ISA) of available microcontrollers can be a great limitation to running quantized neural networks. Indeed, most microcontroller architectures do not exhibit any kind of SIMD instructions. On the other hand, most microcontrollers rely on 32-bit registers. Thus, even if the neural network parameters and the input data use a lower precision representation, they have to be computed one by one using the register width of 32 bits.
Some more advanced microcontroller architectures offer instructions able to handle 4 × 8-bit or \mbox{2 × 16-bit} data packed in 32-bit registers. However, such advances do not allow working with intermediate precision or sub-byte precision, and not all arithmetic and logic instructions are covered.
Intermediate or sub-byte precision requires manually packing and unpacking data, thus inducing a noticeable computation overhead, even though it helps further reducing the memory footprint.

Moreover, microcontrollers used in IoT devices mostly rely on ARM Cortex-M cores with the associated ISA. However, ARM cores are not open, meaning that modifying and extending the ISA is not possible.
To overcome these limitations, the authors in \cite{XpulpNN} proposed an extension to the RISC-V ISA, which is open, with instructions to handle sub-byte quantization.
Unfortunately, as microcontrollers implementing RISC-V are still scarce on the market, and not readily available with the proposed extension, this approach cannot be reasonably used to deploy IoT devices since it requires manufacturing a custom microcontroller. Manufacturing a custom microcontroller is not feasible when the goal is to release an IoT product on the market, due to large costs, time and the required level of expertise.
As a result, only off-the-shelf microcontrollers are considered in this work. Only 8-bit, 16-bit and 32-bit precision will therefore be studied.

Deep neural networks have already been deployed on 8-bit microcontrollers. One of the first methods was proposed in \cite{NN8BitMCU}. Although interesting, this method requires a lot of work to implement pseudo-floating-point coding, a custom multiplication algorithm over 16 bits, as well as a hyperbolic tangent approximation for the activation function, all in assembly language.
Over the last few years, implementations have relied on 32-bit microcontrollers with either a hardware FPU or fixed-point computations.
In addition, the Rectified Linear Unit (ReLU)~\cite{ReLu} has become widely used as an activation function and has the benefit of being easily computed as a max between 0 and the layer's output, thus being much less complex than a hyperbolic tangent.
In the meantime, neural network architectures and training methods have continued to evolve to provide more and more efficient models.
As a result, applications such as spoken keyword spotting~\cite{KeywordSpotting} and human activity recognition~\cite{DSD2020} can now be performed in real time on IoT devices relying on low-power microcontrollers.

\section{Representation of Real Numbers}
\label{sec:realnumbers}
\subsection{Floating-Point}
In modern computation systems, real numbers typically use a floating-point representation.
Floating-point representation relies on the encoding of three different pieces of information: the sign, the significand and the exponent. Coding the significand and the exponent separately makes it possible to represent values with a very large dynamic range, while at the same time providing increasing precision as the numbers approach 0.

Most floating-point implementations follow the IEEE754~\cite{IEEE754} standard which defines how the sign, significand and exponent are coded in a binary format.
Floating-point numbers can be coded in half, single or double precision requiring 16, 32 or 64 bits, respectively.
Obviously, the more bits allocated to code a value, the more precise it is. A higher number of bits allocated to the exponent also allows for a larger dynamic range.
In deep neural networks, single precision is more than enough for training and inference.
Double precision requires more computing resources, so it is generally not used.
Recently, it has been shown that half-precision can further accelerate the training and inference without a significant drop in the accuracy of the deep neural network~\cite{MixedPrecision}.

However, the choice is much more restricted for low-power microcontrollers. When present, the hardware floating-point unit often only supports single-precision computation. Double-precision computations must be performed in software and are therefore significantly slower than single precision. Half-precision data are converted to single precision before the computation. In 2019, ARM released the ARMv8.1-M ISA which includes instructions for half-precision support. Even though the Cortex-M55 core is planned to implement these instructions, there is so far no microcontroller with this core available on the market.
As a result, when floating point is used on a microcontroller, only single precision is~considered.

The binary representation of single-precision floating-point numbers is called binary32 and is represented with 1 bit for the sign, 8 bits for the exponent and 23 bits for the significand (see Table \ref{table:fp32}).
It allows a dynamic range of roughly $[-10^{38}, 10^{38}]$, far beyond the values typically seen in a deep neural network, while increasing the resolution for numbers close to $0$. The closest possible numbers to $0$ are approximately $\pm 1.4\times 10^{-45}$.

\begin{specialtable}[H]
	\caption{Single-precision floating-point binary32 representation.} 
	\resizebox{\columnwidth}{!}{
	\begin{tabular}{p{1.3em}p{0em} p{0em} p{0em} p{0em} p{0em} p{0em} p{0em} p{0.8em}  p{0em} p{0em} p{0em} p{0em} p{0em} p{0em} p{0em} p{0em} p{0em} p{0em} p{0em} p{0em} p{0em} p{0em} p{0em} p{0em} p{0em} p{0em} p{0em} p{0em} p{0em} p{0em} p{0.4em} }
	\toprule
	31 & 30 & 29 & 28 & 27 & 26 & 25 & 24 & 23 & 22 & 21 & 20 & 19 & 18 & 17 & 16 & 15 & 14 & 13 & 12 & 11 & 10 & 9 & 8 & 7 & 6 & 5 & 4 & 3 & 2 & 1 & 0 \\
	\midrule
	sign & \multicolumn{8}{c}{exponent} & \multicolumn{23}{c}{significand} \\
	\bottomrule
	\end{tabular}
	}
	\label{table:fp32}
\end{specialtable}

\subsection{Fixed-Point}

Fixed-point is another way to represent real numbers. In this representation, the integer part and the fractional part have a fixed length.
As a result, the dynamic range and the resolution are directly limited by the length of the integer part and the length of the fractional part, respectively. The resolution is constant across the whole dynamic range.
In binary, the $Q$ notation is often used to specify the number of bits associated with each part. $Qm.n$ is a number where $m$ bits are allocated to the integer part and $n$ bits to the fractional part~\cite{Qformat}.
It is important to note that we consider signed numbers in two's complement representation, the sign bit being included in the integer part.
The number of bits for the integer part can be increased to obtain a larger dynamic range, but it will conversely reduce the number of bits allocated to the fractional part, thus reducing its precision.

Given a $Qm.n$ signed number, its dynamic range is $[-2^{m-1}, 2^{m-1} - 2^{-n}]$ and its resolution is $2^{-n}$.

As an example, in Table \ref{table:fixed32}, a signed $Q16.16$ number stored in a 32-bit register has \mbox{16 bits} for the integer part including 1 bit for the sign and 16 bits for the fractional part.
This translates to a dynamic range of $[-32,768,\,32,767.9999847]$, much smaller than the equivalent floating-point representation, and a constant resolution of $1.5259 \times 10^{-5}$ across the whole range, less precise than the floating-point representation near $0$.

\begin{specialtable}[H]
	\caption{Fixed-point Q16.16 on 32-bit representation.}
	\resizebox{\columnwidth}{!}{
	\begin{tabular}{p{0em} p{0em} p{0em} p{0em} p{0em} p{0em} p{0em} p{0em} p{0em} p{0em} p{0em} p{0em} p{0em} p{0em} p{0em} p{0.8em}  p{0em} p{0em} p{0em} p{0em} p{0em} p{0em} p{0em} p{0em} p{0em} p{0em} p{0em} p{0em} p{0em} p{0em} p{0em} p{0.4em} }
	\toprule
	31 & 30 & 29 & 28 & 27 & 26 & 25 & 24 & 23 & 22 & 21 & 20 & 19 & 18 & 17 & 16 & 15 & 14 & 13 & 12 & 11 & 10 & 9 & 8 & 7 & 6 & 5 & 4 & 3 & 2 & 1 & 0 \\
	\midrule
	\multicolumn{16}{c}{integer part} & \multicolumn{16}{c}{fractional part} \\
	\bottomrule
	\end{tabular}
	}
	\label{table:fixed32}
\end{specialtable}

\section{Training and Quantization of Deep Neural Networks}
\label{sec:nnquant}
In this work, the training is always performed using single-precision floating-point computation, that is, in the \textit{binary32} format.
As training is done offline on a workstation, there is no need to perform it in fixed point. Despite this being feasible, it would come with additional challenges regarding gradient computation.

\subsection{Floating-Point to Fixed-Point Quantization of a Deep Neural Network}
\label{subsec:fp_to_fix}
Since the training relies on floating-point computation, a conversion from a floating-point to a fixed-point representation must be performed before the deep neural network is deployed on the target. As the set of possible values is different between floating-point and fixed-point representations, this involves quantizing the weights of the deep neural~network.

Floating-point to fixed-point conversion requires determining a scale factor, so that the floating-point number can be represented as an integer multiplied by a scale factor. The scale factor is a positive or negative power of two so that it can be computed using only left or right shifts. In the case of the Cortex-M4 architecture, both multiplication and shift instructions take one cycle. However, a division requires 2 to 12 cycles. Therefore, divisions should be avoided as much as possible.

The scale factor must be chosen to represent the whole range of values while avoiding any risk of data overflow, but at the cost of a lower precision for smaller numbers.

\subsubsection{Uniform and Non-Uniform}
\begin{figure}[H]
	\includegraphics[width=0.95\linewidth]{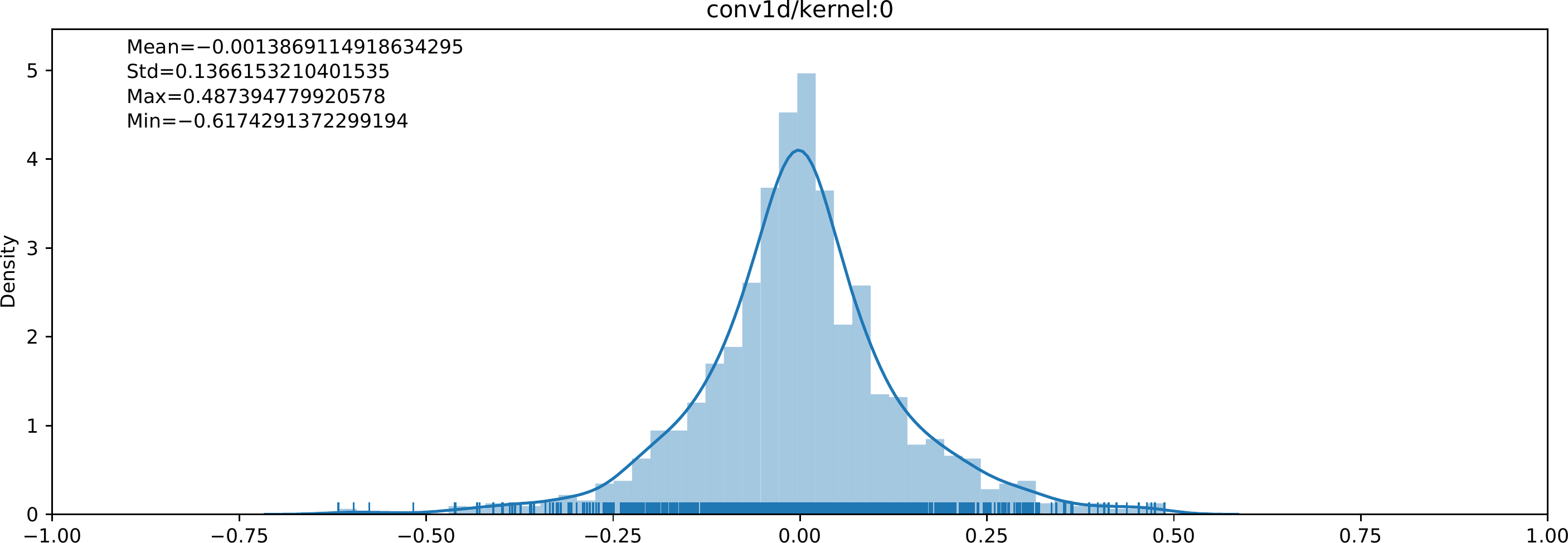}
	\caption{Example of the distribution of weights for a convolutional layer kernel.} 
	\label{fig:distriweightsconv}
\end{figure}

Similarly to the works presented in Section \ref{sec:sota}, in our experiments we also observed that convolutional layer weights are close to Gaussian distributions, with a mean close to 0 when inputs are normalized. Such a distribution of weights for a convolutional layer kernel is shown in Figure~\ref{fig:distriweightsconv}.
As a result, convolutional layer weights can be better represented using a non-uniform distribution of numbers. This is what floating-point numbers originally do to get a better precision around 0.

However, as the goal is to perform fast computations, a uniform quantization is preferred.
Non-uniform quantization would require performing additional transformations before using the microcontroller's instructions. This overhead can be non-negligible and lead to quantization performance lower than floating-point computations. To obtain a nonconstant quantization step, a nonlinear function must be computed, either online or offline, to generate a lookup table where operands for each operation are stored.
In contrast, uniform quantization is based on a constant quantization step.
Furthermore, coding only the nearest power of two does not bring an improvement on Cortex-M4-based microcontrollers. Multiplications and shifts are implemented in hardware and take only 1 cycle. In consequence, the benefits of this kind of approach are limited.
For these reasons, we will rely on uniform quantization in this work.

\subsubsection{Scaling and Offset}
An alternative consists of finding a scale factor that is not necessarily a power of two, but which scales values in $[-1;+1[$. 
Using this technique, all the bits (except the sign bit) are used to represent the fractional parts of numbers. As an example, there is the $Q1.15$ format for 16-bit numbers. 

The scale factor also uses a fixed-point representation but with its own power of two scale factor.
This allows for a slightly lower quantization error since the quantization step is not necessarily a power of two.
However, scaling a number adds computations: instead of performing only a shift, a multiplication with the scale factor followed by a shift with the scale factor's power-of-two scale factor must be performed.

In addition, 
the range could be made asymmetric by introducing an offset value. This would also enable a slightly lower quantization error when the distribution is not centered around 0. However, this requires one more addition when scaling a number.
It is worth noting that using unsigned numbers for ReLU activation could help recovering one more bit for the fractional part if fully merged with the previous layer. Nevertheless, it also comes with an additional implementation complexity to perform operations between signed and unsigned numbers.
All these alternatives imply a higher complexity and additional computations, with only a slight improvement of the quantization error. For these reasons, we decided to use a scale factor that is a power of two and a symmetric~range.

\subsubsection{Per-Network, Per-Layer and Per-Filter Scale Factor}
To reach the best quantization, the scale factor should in theory be chosen for each weight value. However, storing a scale factor for each value is obviously not a reasonable approach since it leads to an important memory overhead, defeating the purpose of implementing quantization to reduce memory usage.
On the other hand, using a scale factor for the whole network is too coarse to achieve a good overall quantization error. Instead, the scale factor can be made different for each layer.
Another solution consists of using a scale factor for each filter of each layer. Although more complex to implement and introducing some overhead (scale factors of the layers must be stored in memory), this approach can slightly decrease the quantization error. 
So far, our implementation only allows a per-network and a per-layer scale factor. The parameters and the activations can have different scale factors.

\subsubsection{Conversion Method}
\label{subsubsec:conversionmethod}
To convert from floating-point to fixed-point, the method starts with finding the required number of bits $m$ to represent the  {unsigned} integer part:

\begin{equation}
m = 1 + \left \lfloor{\log_2(\max_{1\le i\le N}|x_i|)}\right \rfloor
\end{equation}
where $x_i$ is an element of the floating-point vector $x$ of length $N$.
A positive value of $m$ means that $m$ bits are required to represent the absolute value of the integer part, while a negative value of $m$ means that the fractional part has $m$ leading unused bits. This enables a greater precision to be obtained for vectors with numbers smaller than $2^{-1}$, since the leading unused bits can be removed and replaced instead by more trailing bits for precision.

From this we can compute the number of remaining bits $n$ for the fractional part:
\begin{equation}
n = w - m - 1
\end{equation}
where $w$ is the data type width (e.g., 16 bits).

In this equation, 1 is subtracted to take into account the additional bit required to represent signed numbers.

A positive value of $n$ means that $n$ bits are available to represent the fractional part.

A negative value of $n$ means that the fractional part cannot be represented, and the integer part cannot be represented to its full precision.

An element $xfixed_i$ of the fixed-point vector $xfixed$ is computed from the element $x_i$ of the floating-point vector $x$ as:
\begin{equation}
xfixed_i = floor(x_i \times 2^n)
\end{equation}
where $trunc(y)$ truncates a real number $y$ to its integer part.

And the scale factor $s$ is defined as:
\begin{equation}
s = 2^{-n}
\end{equation}

Two methods can be used to get the quantized weights of a deep neural network. These methods are detailed in the following.

\subsection{Post-Training Quantization}
In post-training quantization, the neural network is entirely trained using floating-point computation (a \textit{binary32} format is assumed here). Once the training is over, the neural network is frozen, and the parameters are then quantized. The quantized neural network is then used to perform the inference, without any adjustments of the parameters.

The quantization phase introduces a quantization error on each parameter as well as on the input, thus leading to a quantization error on the activations. The accumulation of quantization errors at the output of the neural network can cause the classifier to incorrectly predict the class of the input data, creating an accuracy drop compared to the non-quantized version. As the bit width of the values decreases, the quantization error increases, and the resulting accuracy typically decreases as well. In some situations, a slight increase in the quantization error can help the network generalize better over new data, inducing a slight increase in the accuracy over test data.

\subsection{Quantization-Aware Training}
\label{subsec:qat}
The objective of the quantization-aware training (QAT) is to compensate the quantization error by training the deep neural network using the quantized version during the forward pass. This should help to mitigate the accuracy drop to some extent.
The backpropagation still relies on non-quantized values.
To stabilize the learning phase with the quantized version, and thus obtain better results on average, the DNN can be pre-trained using a floating-point representation in order to initialize the parameters to sensible~values.

\begin{figure}[H]
\includegraphics[width=0.95\linewidth]{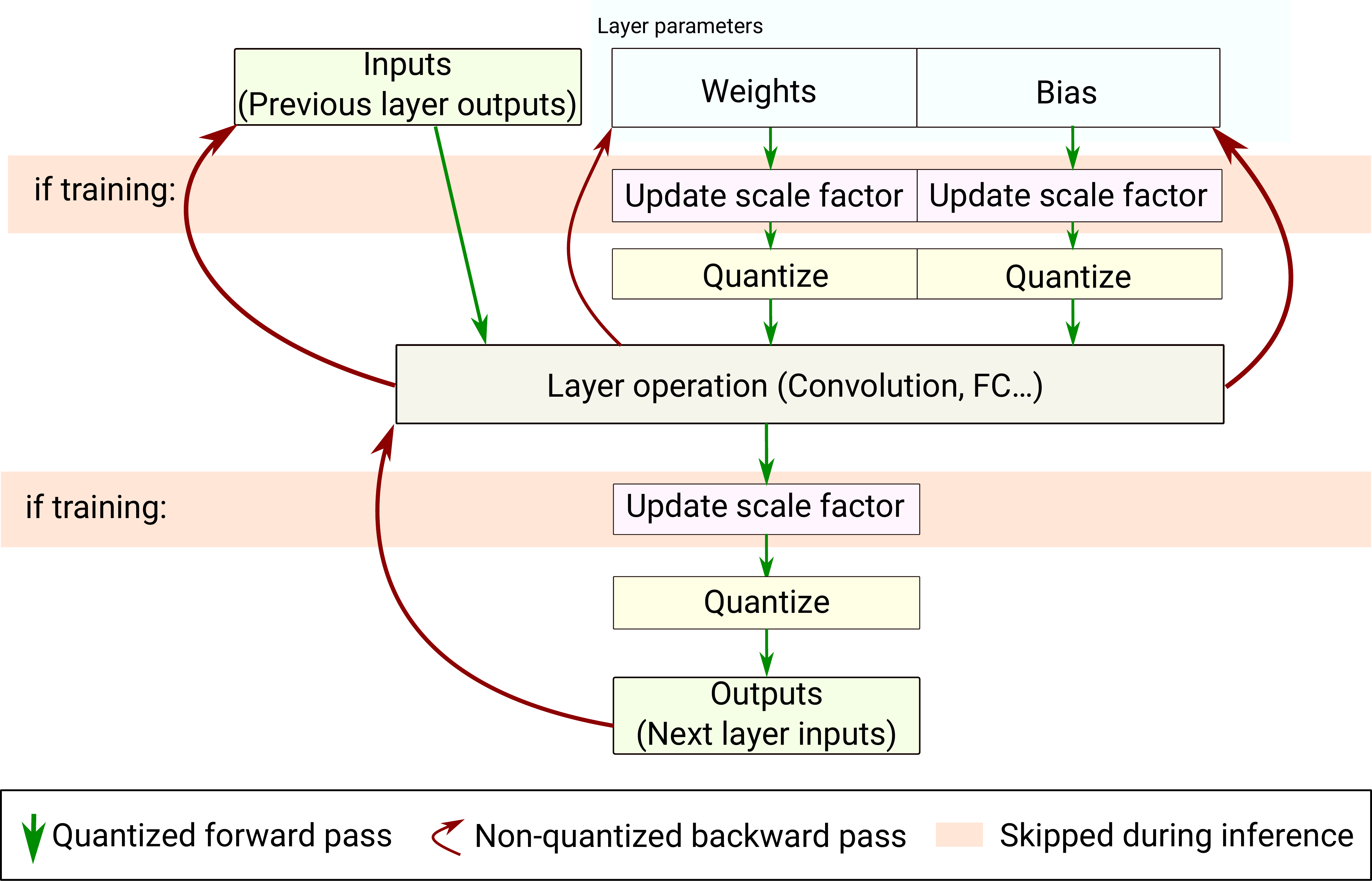}
\caption{Quantization-aware training}
\label{fig:training_flow}
\end{figure}

In this work we decided to perform all the computations using a floating-point representation. As shown in Figure~\ref{fig:training_flow}, the inputs, weights and biases of each layer are quantized (but kept in floating-point representation) before actually performing the layer's computation. The layer's output is quantized after the computation, before reaching the next layer.
The quantization operation is done following the method presented in \mbox{Section \ref{subsec:fp_to_fix}}.
During the training phase, the range of values is reassessed each time to adjust the scale factor before performing the layer's computation. When doing inference only, the scale factor is frozen.

In case of a convolutional neural network, the convolutional and fully connected layers require a quantization-aware training for the weights.
Please note that batch normalization layers also require quantization-aware training. However, as we do not use batch normalization in our experiments, it has not been implemented.
For max-pooling layers, quantization-aware-training is not required as they do not have weights.
Moreover, as max-pooling only consists of an element-wise max, there is no need to quantize: inputs are already quantized from the previous layer and the dynamic range cannot be expanded.
Therefore, no quantization is done on the max-pooling layers.
It is similar for the ReLU activation which is considered to be a separate layer.
Conversely, the element-wise addition layer requires quantization. It does not have trainable weights; however, the dynamic range of the output can increase after adding two large numbers. Therefore, the same quantization process is applied to compute the output scale factor.

\section{Deployment of the Quantized Neural Network}
\label{sec:deployment}
After the network has been trained and optionally quantized, it is deployed onto a microcontroller to perform the inference on the target platform.
Deployment involves the following phases:
\begin{itemize}
    \item exporting the weights of the deep neural network and encoding them into a format suitable for on-target inference, 
    \item \textls[-5]{generating the inference program according to the topology of the deep neural~network,} 
    \item compiling the inference program, and
    \item uploading the inference program with the weights onto the microcontroller's~ROM.
\end{itemize}

\subsection{Existing Embedded AI Frameworks}
Several embedded AI frameworks are already available. Among them, the most popular ones are TensorFlow Lite for Microcontrollers~\cite{TFLiteMicro} and STM32Cube.AI~\cite{STM32CubeAI}. Other frameworks stemming from research projects also exist  and are discussed in the following.

\subsubsection{TensorFlow Lite for Microcontrollers}
TensorFlow Lite for Microcontrollers (or TFLite Micro) is a project derived from TensorFlow Lite. Originally focused on deep neural network deployment on smartphones, it has been made available for microcontrollers.
TFLite Micro supports a wide range of operations~\cite{TFLiteMicroOps}, enabling the deployment of a variety of deep neural networks, such as multi-layer perceptrons and convolutional neural networks, including residual neural networks.
Deep neural networks are developed and trained using TensorFlow/Keras, and can then be deployed semi-automatically onto a microcontroller.

TFLite Micro is intended to be generic enough to be deployed on any kind of \mbox{32-bit} microcontroller. The inference library is therefore portable, but it also means there is no integration with specific microcontroller families and vendor tools.
The trained deep neural network (topology and weights) can be automatically converted to a format understandable by the inference library, but there are no tools to generate and deploy the application code.
Moreover, the test application must be written by hand. 
Nevertheless, a template source code for a few development boards (e.g., the SparkFun Edge) as well as a few demo applications (e.g., keyword spotting) are available.
Finally, TFLite Micro does not come with tools to measure metrics such as the inference time or the RAM and ROM~usage.

TFLite Micro supports computation in both floating point in \textit{binary32} format and fixed point on 8-bit integers. The quantization technique uses a non-power-of-two scale factor, a symmetric range for the weights and an asymmetric range for the activations. Biases are quantized on 32-bit integers.
Convolution operations can make use of a per-filter scale factor and offset, while other operations use a per-tensor (i.e., per-layer) scale factor and offset~\cite{TFLiteQuantSpec,TFQuantInf}.
There is no support for fixed point on 16-bit integers.

Inference with 8-bit integers can be accelerated using low-level optimizations provided by the CMSIS-NN \cite{CMSISNN} library from ARM. This library uses SIMD-like instructions (from the ARMv7E-M instruction set architecture of Cortex-M4 cores) to perform two multiply–accumulate (MACC) operations on 16-bit operands with a single 32-bit accumulator in one cycle.

While being entirely free/open-source, the complexity of the software architecture makes it quite difficult to manipulate and extend. This is a substantial drawback in a research environment, and it also comes with additional overhead.
The deep neural network topology is deployed as a sort of microcode that is interpreted at runtime instead of being statically compiled. This process makes it more difficult for the compiler to perform optimizations {and causes} a larger memory usage.

\subsubsection{STM32Cube.AI}
STM32Cube.AI is a software suite from STMicroelectronics that enables the deployment of deep neural networks onto their STM32 family of microcontrollers.
STM32Cube.AI supports deployment of trained deep neural network models from several frameworks including Keras and TensorFlow Lite.
A wide range of operations are supported~\cite{STM32CubeAISupport}, allowing the deployment of several deep neural network architectures, such as multi-layer perceptron and convolutional neural networks, including residual neural networks.

STM32Cube.AI is a software suite fully integrated with other STMicroelectronics development and deployment tools such as STM32CubeMX, STM32CubeIDE and \linebreak STM32CubeProgrammer. This provides a very straightforward and easy to use flow. Moreover, a test application is included to evaluate the model on target with a real test dataset, providing metrics on inference time, and ROM and RAM usage, without having to write a single line of code.

Like TFLite Micro, STM32Cube.AI supports computations in floating-point \textit{binary32} format and fixed point on 8-bit integers. In fact, the quantization on 8-bit integers comes from TFLite. There is no support for fixed point on 16-bit integers.

STM32Cube.AI also has an optimized inference engine that seems to be partially based on CMSIS-NN. However, as the source code of the inference engine is not freely available, it is not clear what is optimized and how.

The inference library is entirely proprietary/closed-source, therefore it is not possible to manipulate and extend this library. This represents a major drawback in a research environment. It is also not possible to use STM32Cube.AI on microcontrollers which are not part of the STMicroelectronics portfolio.
The inference process and optimizations are not detailed, but unlike TFLite Micro, the network topology is compiled into a set of function calls to the closed-source library rather than being interpreted at runtime.

\subsubsection{Other Frameworks}
Some other frameworks have been developed as part of research projects. These frameworks mainly focus on ``classical'' machine learning (SVM, Decision Tree, etc.), for example, emlearn~\cite{emlearn} and Micro-LM~\cite{MLonMainstreamMCU}, or multi-layer perceptron, for example, Gravity~\cite{Gravity} and FANN-on-MCU~\cite{FANNonMCU}. These frameworks do not support convolutional neural networks with residual connections.
At the time of this work, microTVM seems less mature and popular than TensorFlow Lite for Microcontrollers and STM32Cube.AI. It is, therefore, not studied in this work.

\subsection{MicroAI: our Framework Proposition}
As mentioned above, existing tools for quantized neural networks have some drawbacks that motivated the development of our own framework. This framework addresses the following issues:
\pagebreak
\begin{itemize}
	\item open-source frameworks do not support convolutional neural networks with non-sequential topologies,
	\item frameworks that support convolutional neural networks are proprietary or too complex to be modified and extended easily,
	\item other frameworks do not provide 16-bit quantization,
	\item some frameworks are dedicated to a limited family of hardware targets.
\end{itemize}

In this work, we aim at providing a framework that is easy to extend and modify, and that allows for a complete pipeline from the neural network training to the deployment and evaluation on the microcontroller.
Additionally, this framework must provide a lightweight runtime on the microcontroller to reduce the overhead.
Finally, our objective is to achieve a performance close to existing solutions.

Our framework, called MicroAI, is built in two parts:
\begin{enumerate}
	\item A neural network training code that relies on Keras or PyTorch.
	\item A conversion tool (called KerasCNN2C) that takes a trained Keras model and produces a portable C code for the inference
\end{enumerate}

Both parts are written in Python since it is the most popular programming language to build deep neural networks and it easily interfaces with existing frameworks, libraries and tools.

\subsection{MicroAI: General Flow}

\begin{figure}[H]
	\includegraphics[width=0.605\linewidth]{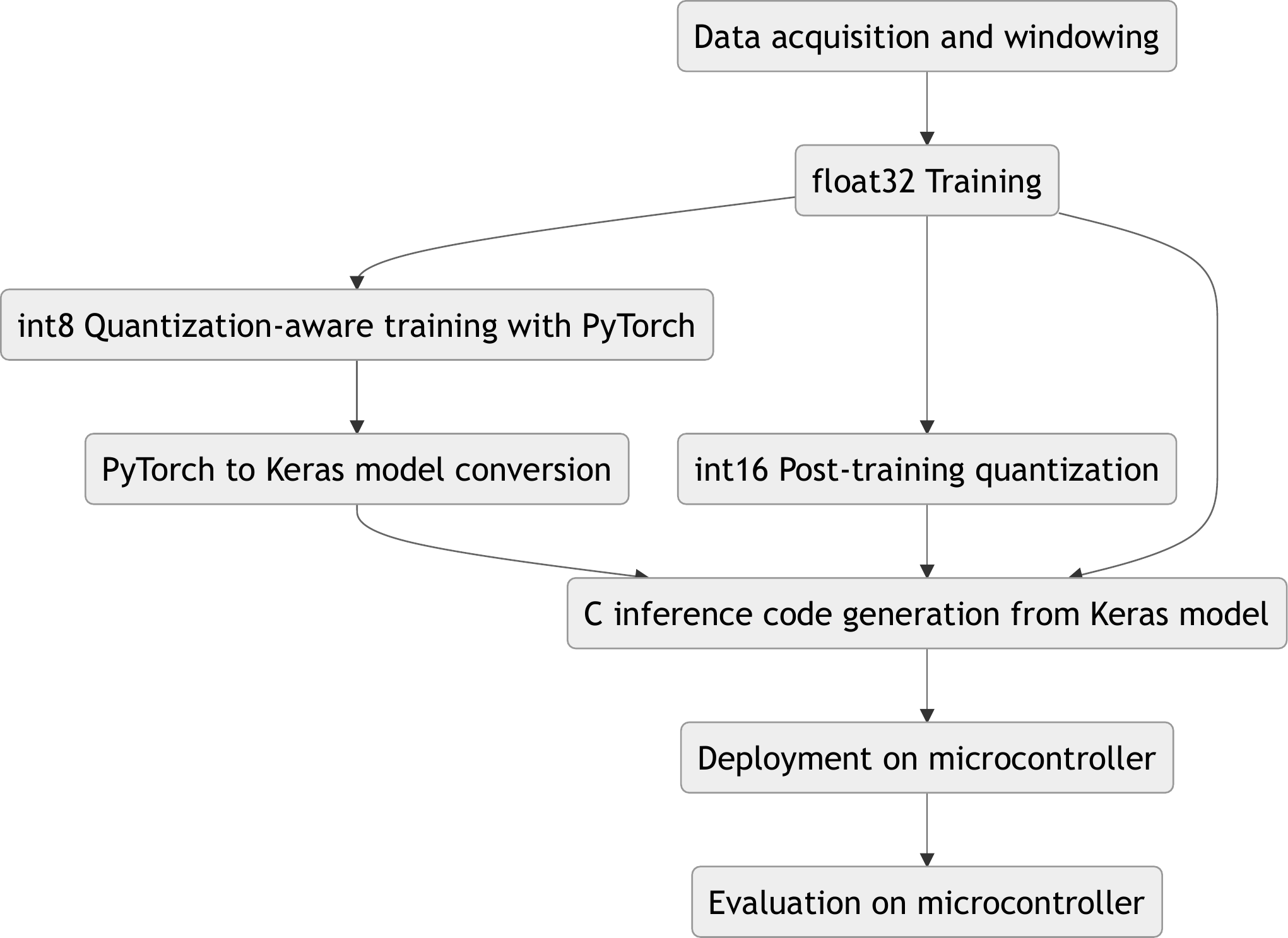}
	\caption{MicroAI general flow for neural network quantization and evaluation on embedded target.}
	\label{fig:microaiflow}
\end{figure}

As seen in Figure~\ref{fig:microaiflow}, MicroAI provides an interface to automatically train, deploy and evaluate an artificial neural network model.
A configuration file written in TOML~\cite{TOML} is used to describe the whole flow of an experiment:
\begin{itemize}
	\item The number of iterations for the experiment (for statistical purposes)
	\item The dataset to use for training and evaluation
	\item The preprocessing steps to apply to the dataset
	\item The framework used for training
	\item The various model configurations to train, deploy and evaluate
	\item The configuration of the optimizer
	\item The post-processing steps to apply to the trained model
	\item The target configuration for deployment and evaluation
\end{itemize}

The three main steps, training, deployment and evaluation, are described in the following. The commands used to trigger them are available in Appendix \ref{appendix:microaicommands}.

\subsection{MicroAI: Training}
For the training phase, MicroAI is simply a wrapper around Keras or PyTorch.

A dataset requires an importation module to be loaded into an appropriate data model.
The training process expects a \texttt{RawDataModel} instance, which gathers the training and test sets. This instance contains \textit{numpy} arrays for the data and the labels.
A higher-level data model \texttt{HARDataModel} is also available for human activity recognition to process subjects and activities more easily. This model is then converted to a \texttt{RawDataModel} using the \texttt{DatamodelConverter} in the preprocessing phase.
The preprocessing phase also includes features such as normalization.
Dataset importation modules for UCI-HAR, SMNIST and GTSRB (described in Section \ref{sec:results}) are included and can easily be extended to new datasets.

To make use of a deep neural network architecture in the model configuration, it must be first described according to the training framework API in use.
The description of the model is a template where parameters can be set by the user in the configuration file.
MicroAI provides the following neural network architectures for both Keras and PyTorch:

\begin{itemize}
	\item MLP: a simple multi-layer perceptron with a configurable number of layers and neurons per layer.
	\item CNN: a 1D or 2D convolutional neural network with configurable number of layers, filters per layer, kernel and pool size, and number of neurons per fully connected layer for the classifier 
	\item ResNet: a 1D or 2D residual neural network (v1) with convolutional layers. The number of blocks and filters per layer, stride, kernel size, and optional BatchNorm can be configured. 
\end{itemize}

All architectures use ReLU activation units.

In the configuration file, several model settings can be described, each inside their own \texttt{[[model]]} block. Each model will be trained sequentially.
A common configuration for all the models can be specified in a \texttt{[model\_template]} block. Model configuration also includes optimizer configuration and other parameters such as the batch size and the number of epochs.

Once the model is trained, some post-processing can be applied. It is for instance possible to remove the SoftMax layer for Keras models with the \texttt{RemoveKerasSoftmax} module. This layer is indeed useless when only inference is performed.

{Even though it also performs model training, the quantization-aware training described in Section \ref{subsec:qat} is also included {for PyTorch} as a post-processing step in the} \linebreak \texttt{QuantizationAwareTraining} module.
The actual training step before post-processing is seen as a general training, before optionally performing post-training quantization or quantization-aware training.
The quantization-aware training can be seen as a fine-tuning on top of the more general training (which can also be skipped if necessary).
The quantization-aware training does not actually convert the weights from a floating-point data type to an integer data type with fixed-point representation. This conversion is instead performed by the KerasCNN2C conversion tool.

\textls[-10]{Support for additional learning frameworks can be added by creating a new class implementing the \texttt{LearningFramework} interface and by supplying compatible model templates.}

\subsection{MicroAI: Deployment}
MicroAI can deploy a trained model to perform inference on a target using either STM32Cube.AI, TensorFlow Lite for Microcontrollers or our own tool, KerasCNN2C.

STM32Cube.AI can be used for all STM32 platforms, and support for the Nucleo-L452RE-P with an STM32L452RE microcontroller is included.
Support for other platforms using a STM32 microcontroller can be added by providing a sample STM32CubeIDE project including the X-CUBE-AI package.
STM32Cube.AI does not support microcontrollers outside the STM32 family.

TensorFlow Lite for Microcontrollers is a portable library that can be included in any project. Therefore, it could be used for any 32-bit microcontroller. However only integration with the SparkFun Edge platform with an Ambiq Apollo3 microcontroller is included in our framework so far.

Similarly, KerasCNN2C produces a portable library that can be included in any project. So far, only integration with the Nucleo-L452RE-P and the SparkFun Edge boards has been performed. Support for other platforms can be added by providing project files that call the inference code and a module that interfaces with the build and deployment tools for that~platform.

Please note that none of these tools can take a trained PyTorch model as an input to deploy onto a microcontroller.
The trained PyTorch model must therefore be converted to a Keras model prior to the deployment.
Our framework provides a module to perform semi-automatic conversion from a PyTorch model to a Keras model. A Keras model that matches the structure of the PyTorch model must be programmed and the matching between the PyTorch model and Keras model layer names also must be specified. The semi-automatic conversion module can then automatically copy the weights from the PyTorch model to the Keras model and export it for use by one of the deployment tools.

\subsection{KerasCNN2C: Conversion Tool from Trained Keras Model to Portable C Code}
KerasCNN2C is a tool that we developed to automatically generate, from a trained Keras model exported as an \texttt{HDF5} file, a C library for  inference. It can also be used independently of the MicroAI framework.

 In this work, only 1D models are evaluated on target. Work is underway for full support of 2D-model deployment. Training and quantization are already supported, therefore 2D models are evaluated offline. Here are the supported layers so far:
\begin{itemize}
	\item Add
	\item AveragePooling1D
	\item BatchNormalization
	\item Conv1D
	\item Dense
	\item Flatten
	\item MaxPooling1D
	\item ReLU
	\item SoftMax
	\item ZeroPadding1D
\end{itemize}

Layers can have multiple inputs such as the \textit{Add} layer, thus allowing residual neural networks (ResNet) to be built.
Sequential convolutional neural networks or multi-layer perceptron models are also supported.

The generated library exposes a function in the \texttt{model.h} header to run the inference process with the following signature:

\begin{lstlisting}
void cnn(
  const number_t input[MODEL_INPUT_CHANNELS][MODEL_INPUT_SAMPLES],
  output_layer_type output);
\end{lstlisting}
\vspace*{-18pt}
\noindent where \texttt{number\_t} is the data type used during inference defined in the \texttt{number.h} header, and \texttt{MODEL\_INPUT\_CHANNELS} and \texttt{MODEL\_INPUT\_SAMPLES} are the dimensions of the input defined in the generated \texttt{model.h} header.
The input and output arrays must be allocated by the caller.

The model inference function does not proceed to the conversion of the input from fixed-point to floating-point representation when using a fixed-point inference code.
The caller must perform the conversion before feeding the buffer to the model inference function (see Section \ref{sec:microaiquantization}).

To this aim, a floating-point number \texttt{x\_float} can be converted to a fixed-point number \texttt{x\_fixed} with the following call:

\begin{lstlisting}
x_fixed = clamp_to_number_t((long_number_t)floor(x_float*(1<<INPUT_SCALE_FACTOR)));
\end{lstlisting}
\vspace*{-18pt}
\noindent where \texttt{long\_number\_t} is a type twice the size of \texttt{number\_t} and \texttt{clamp\_to\_number\_t} saturates and converts to \texttt{number\_t}. Both are defined in the \texttt{number.h} header.\linebreak
\texttt{INPUT\_SCALE\_FACTOR} is the scale factor for the first layer, defined in the \texttt{model.h} header.

The output array corresponds to the output of the model's last layer, which is typically a fully connected layer when solving a classification problem. If the purpose is to predict a single class, the caller must find the index of the max element in the output array.

\subsection{KerasCNN2C: Conversion Process}
This tool first parses the model using the Keras API from TensorFlow 2.4 and generates an internal representation of the topology (i.e., a graph), with each node corresponding to a layer.

Then, a series of transformations is performed to produce a graph better suited for deployment on a microcontroller:
\begin{itemize}
	\item combine ZeroPadding1D layers (if they exist) with the next Conv1D layer,
	\item combine ReLU activation layers with the previous Conv1D, MaxPooling1D, Dense or Add layer,
	\item convert BatchNorm{\cite{batchnorm}} weights from the mean $\mu$, the variance $V$, the scale $\gamma$, the offsets $\beta$ and $\epsilon$ to a multiplicand $w$ and an addend $b$ using the following formula: 
	\begin{equation}
		w = \dfrac{\gamma}{\sigma}
	\end{equation}
	\begin{equation}
		\sigma = \sqrt{V + \epsilon}
	\end{equation}
	\begin{equation}
		b = \beta - \dfrac{\gamma \times \mu}{\sigma}
	\end{equation}
	so that the output of the BatchNorm layer can be computed as $y = w\times x + b$. It could be folded in the previous convolutional layer, but this is not implemented yet.
\end{itemize}

Then, for each node in the graph, the weights of the layer go through the quantization and conversion module if the conversion to fixed-point representation is enabled. 
The C inference function is generated from a Jinja2~\cite{Jinja2} template file using the layer's configuration. Similarly, the layer's weights are converted into a C array from a Jinja2 template file.
Code generation is used to avoid runtime overhead of an interpreter such as the one used in TensorFlow Lite for Microcontrollers. Additionally, it allows the compiler to perform better optimizations.
In fact, the layer's configuration is generated as constants or literals in the code, allowing the compiler to perform appropriate optimizations such as loop unrolling, using immediates when relevant and doing better register allocation.
By default, GCC's \texttt{-Ofast} optimization level is enabled.
Moreover, the code is written in a simple and straightforward way.
So far, no special effort has been made to further optimize the source code for faster execution.

The allocator module aims to reduce RAM usage. To do so, it allocates the layer's output buffers in the smallest number of pools without conflicts. For each layer of the model, its output buffer is allocated to the first pool that satisfies two conditions: it must neither overwrite its input, nor the output of a layer that has not already been consumed. If there is no such available pool, a new one is created.
It is worth noting that the allocator module does not yet try to optimize the allocation to minimize the size of each pool (this is a harder problem to solve). In consequence, the total RAM usage is not optimized.

Finally, the main function \texttt{cnn(...)} is generated. {This function only contains the allocation of the buffers done by the allocator module and a sequence of calls to each of the layers' inference functions. The correct input and output buffers are passed to each layer according to the graph of the model.}

\subsection{KerasCNN2C: Quantization and Fixed-Point Computation} \label{sec:microaiquantization}
The post-training quantization is performed by the quantization module itself.
The scale factor for each layer is found according to the method in Section \ref{subsubsec:conversionmethod}, but it can also be specified manually for the whole network.
The fixed-point coding used for all the weights is computed according to this method as well, and the data type is converted from \texttt{float} to an integer data type, such as \texttt{int8\_t} for 8-bit quantization or \texttt{int16\_t} for 16-bit~quantization.

When doing quantization-aware training, the scale factors are found during the training  phase (also according to the method in Section \ref{subsubsec:conversionmethod}). Therefore, the quantization module reuses them. However, the weights are still in floating-point representation since the training phase only relies on floating-point computation. In consequence, the quantization module must perform a data type conversion similar to the one performed for post-training quantization.

Once the model is deployed and running on the target, the fixed-point computation can be done using a regular integer arithmetic and logic unit.
The data type for the input and output of a layer is the same as the one used to store the weights.
To avoid overflows, computation is done using a data type twice the width of the operands' data type.
For example, if the data type of the weights and inputs is \texttt{int16\_t}, then the intermediate results in a layer are computed and stored with an \texttt{int32\_t} data type.
The result is then scaled back to the correct output scale factor before saturating and converting it back to the original operands' data type.

Before performing an addition or a subtraction, operands must be represented with the same number of integer and fractional bits.
This is not required for multiplication, but the number of bits allocated for the fractional part of the result is the sum of the number of bits for the fractional part of the two operands.
Therefore, after a multiplication, the result must be scaled to the required format by shifting the result to the right by the appropriate number of bits.

{In Appendix \ref{appendix:aluintops}, the number of operations required for the main layers of a residual neural network in our implementation are provided, along with the number of cycles taken for these operations. Enabling compiler optimizations generates some ARMv7E-M instructions, namely \texttt{SMLABB} that performs a multiply–accumulate operation in one cycle (instead of two cycles). However, the compiler does not make use of the \texttt{SSAT} operation that could allow saturating in one cycle. Instead, it uses the same instructions as a regular max operation, that is, a compare instruction and a conditional move instruction requiring a total of two cycles.}

\newpage

\section{Results}
\label{sec:results}

\begin{figure}[H]

	\includegraphics[height=0.6\paperheight]{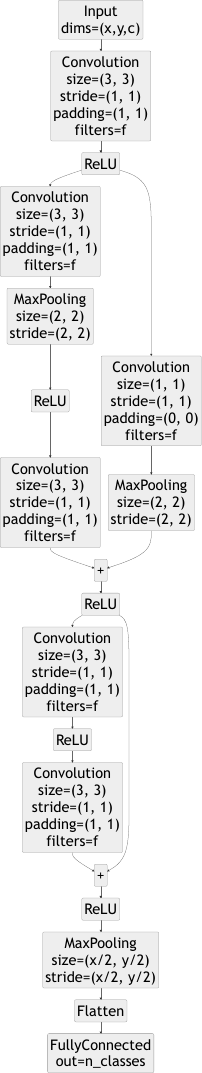}
		\caption{ResNet model architecture.}
	\label{fig:resnet}
\end{figure}

All the results presented in this section rely on the same model architecture, a ResNetv1-6 network with the layers shown in Figure~\ref{fig:resnet}.
The number of filters per layer $f$ is the same for all layers, but is modified to adjust the number of parameters of the model.
The convolutional and pooling layers are one-dimensional except when handling the GTSRB dataset, for which they are two-dimensional.

For each experiment, the residual neural network is initially trained using 32-bit floating-point numbers (i.e., without quantization), and then evaluated over the testing set. This baseline version is depicted as \textit{float32} in the figures shown in the following.

The \textit{float32} neural network is quantized for inference with fixed-point on 16-bit integers and is then evaluated without additional training. This version is depicted as \textit{int16} in the figures shown hereafter.
Quantization is performed using the Q7.9 format for the whole network, meaning the number of bits $n$ for the fractional part is fixed to 9.

The \textit{float32} neural network is also trained and evaluated for inference with fixed-point on 8-bit integers using quantization-aware training. This version is indicated as \textit{int8} in the figures. In this case the fixed-point precision can vary from layer to layer and is determined using the method introduced in Section \ref{subsubsec:conversionmethod}.

The SGD optimizer is used for all experiments. The stability of the SGD optimizer motivated this choice, especially for the quantization-aware training. Training parameters are described below for each dataset. Additionally, training and testing sets are normalized using the z-score of the training set. It is worth noting that Mixup~\cite{Mixup} is also used during~training.

Accuracy is not evaluated directly on the target due to the amount of time it would require. Only inference time for the UCI-HAR dataset is measured on the target.

In the figures, each point represents an average over 15 runs.

\subsection{Evaluation of the MicroAI Quantization Method}
\subsubsection{Human Activity Recognition dataset (UCI-HAR)} 
\label{subsubsec:ucihar}

The University of California Irvine's hosted Human Activity Recognition dataset (UCI-HAR)~\cite{UCIHAR} is a dataset of activities of daily living recorded using the accelerometer and gyroscope sensors of a smartphone.
In this experiment, we use the raw data from the sensors divided into fixed time windows, rather than the precomputed features. The reason is that we want to perform real-time embedded recognition. To do so, it is necessary to avoid the overhead of computing the features for each inference before entering the deep neural network. Instead, the features are extracted by the convolutional neural network itself.

The dataset is divided into a training set of 7352 vectors and a testing set of \mbox{2947 vectors}.
Each vector is a one-dimensional time series of 2.56~s composed of 128 samples sampled at 50~Hz, with 50\% overlap between vectors.
Each sample has 9 channels: 3 axes of total acceleration, 3 axes of angular velocity and 3 axes of body acceleration.
Six different classes are available in the dataset: walking, walking upstairs, walking downstairs, sitting, standing and lying.

The initial training without quantization is performed using a batch size of $64$ over $300$ epochs.
The initial learning rate is set to $0.05$, the momentum is set to $0.9$ and the weight decay is set to $5\times 10^{-4}$.
The learning rate is multiplied by $0.13$ at epochs $100$, $200$ and $250$. The quantization-aware training for fixed-point on 8-bit integers uses the same parameters.

\begin{figure}[H]
	\includegraphics[width=0.9\linewidth,page=1]{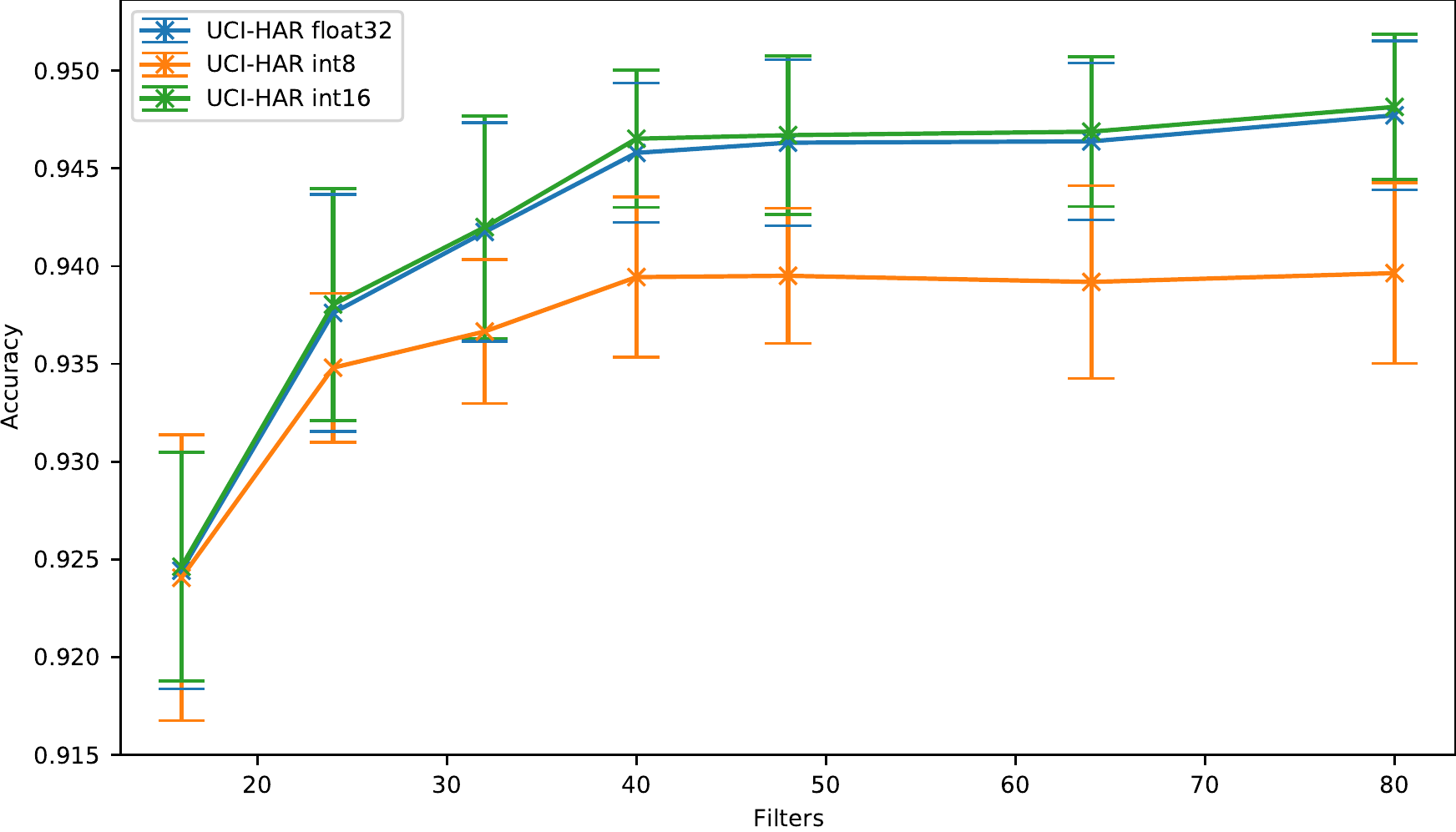}
	\caption{Human Activity Recognition dataset (UCI-HAR): accuracy vs. filters.}
	\label{fig:uciharaccvsfilters}
\end{figure}

As can be seen in Figure~\ref{fig:uciharaccvsfilters}, for the UCI-HAR dataset, the same accuracy is obtained using a 16-bit quantization (\textit{UCI-HAR int16}) or 32-bit floating-point (i.e., the baseline \textit{UCI-HAR float32}), whatever the number of filters per convolution. 

On the other hand, we observe that the 8-bit quantization causes a drop in accuracy that increases in magnitude up to 0.81\% when the number of filters per convolution grows, even though quantization-aware training is used to mitigate this issue.

\begin{figure}[H]
	\includegraphics[width=0.9\linewidth,page=2]{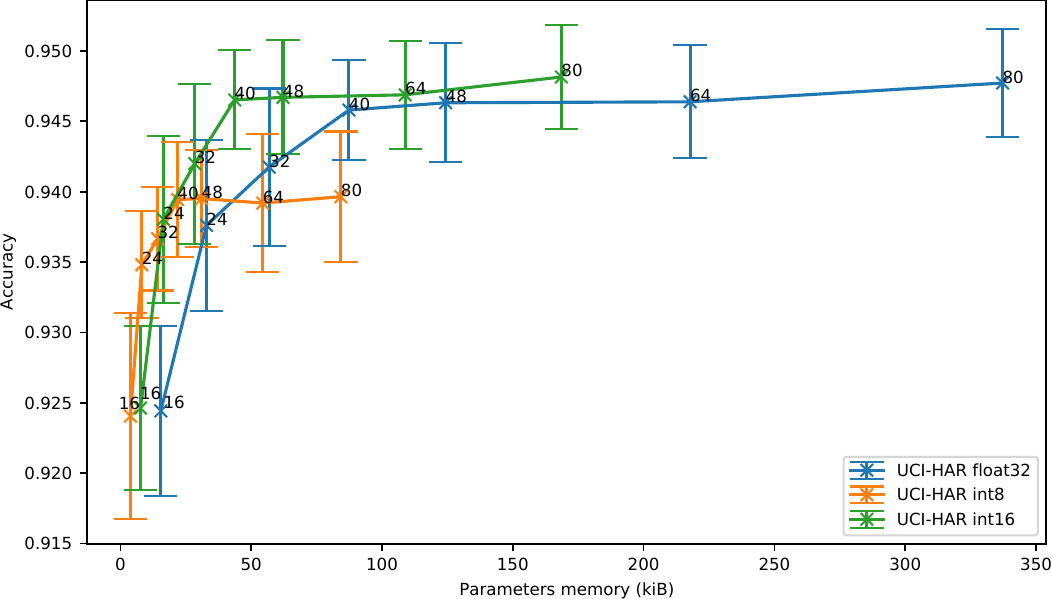}
	\caption{Human Activity Recognition dataset (UCI-HAR): accuracy vs. parameter memory.}
	\label{fig:uciharaccvsparams}
\end{figure}

In Figure~\ref{fig:uciharaccvsparams}, we observe that the accuracy obtained using 8-bit and 16-bit quantization is similar only for deep neural networks exhibiting a reduced number of parameters, in other words a low memory footprint. 
As an example, for 16 filters per convolution, an 8-bit quantization leads to an accuracy of 92.41\% while requiring 3958 memory bytes to store the parameters. When a 16-bit quantization is used, an accuracy of 92.46\% can be achieved, but at the cost of an increase in the required memory for storing the parameters (7916 bytes).

As can be seen, when more than 24 filters per convolution are used, the 16-bit quantization clearly exhibits the best accuracy vs. memory ratio. For more than 48 filters per convolution, the 8-bit quantization provides an even worse ratio than the baseline.

\unskip

\subsubsection{Spoken digits dataset (SMNIST)} 
Spoken MNIST is the spoken digits part of the written and spoken digits database for multi-modal learning \cite{WSMNIST}.

This dataset is made of spoken digits extracted from the Google Speech Commands \cite{GSC} dataset.
The audio signal is preprocessed to obtain \mbox{12 MFCC} plus an energy coefficient using a window of 50~ms with 50\% overlap over the audio files of approximately 1~s each, generating one-dimensional series of 39 samples with \mbox{13 channels}.
The dataset is divided into training and testing sets of 34,801 and \mbox{4107 vectors}, respectively. Some samples are duplicated to obtain 60,000 training vectors and 10,000 testing vectors.
There are 10 different classes for each digit, from 0 to 9.

The initial training, without quantization, uses a batch size of $256$ over $120$ epochs.
The initial learning rate is set to $0.05$, the momentum is set to $0.9$ and the weight decay is set to $5\times 10^{-4}$.
The learning rate is multiplied by $0.1$ at epochs $40$, $80$ and $100$.

The quantization-aware training for fixed-point on 8-bit integers uses a batch size of $1024$ over $140$ epochs.
Initial learning rate, momentum and weight decay are the same as for the initial training.
Learning rate is multiplied by $0.1$ at epochs $40$, $80$, $100$ and $120$.

\begin{figure}[H]
	\includegraphics[width=0.9\linewidth,page=1]{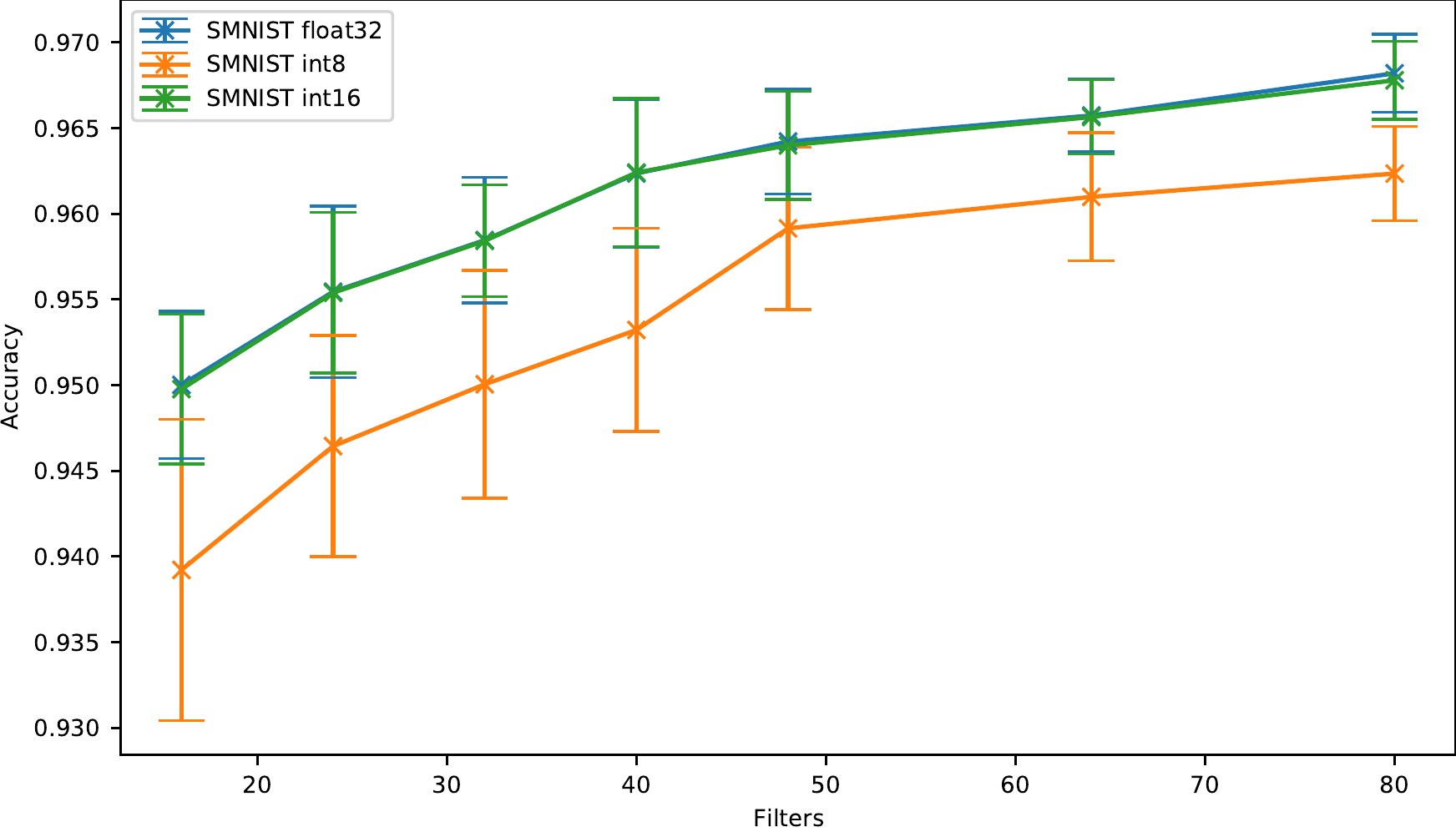}
	\caption{Spoken digits dataset (SMNIST): accuracy vs. filters.}

	\label{fig:smnistaccvsfilters}
\end{figure}

As can be observed in Figure~\ref{fig:smnistaccvsfilters} and regardless of the number of filters, the 16-bit quantization (\textit{SMNIST int16}) provides overall a similar accuracy compared to the floating-point baseline (\textit{SMNIST float32}). On the other hand, {the accuracy drops by up to 1.07\%} when the 8-bit quantization is used. However, the accuracy drop slightly decreases when 48 filters per convolution are used, and then stays around 0.5\% or 0.6\% for a higher number of filters.

\begin{figure}[H]
	
	\includegraphics[width=0.9\linewidth,page=2]{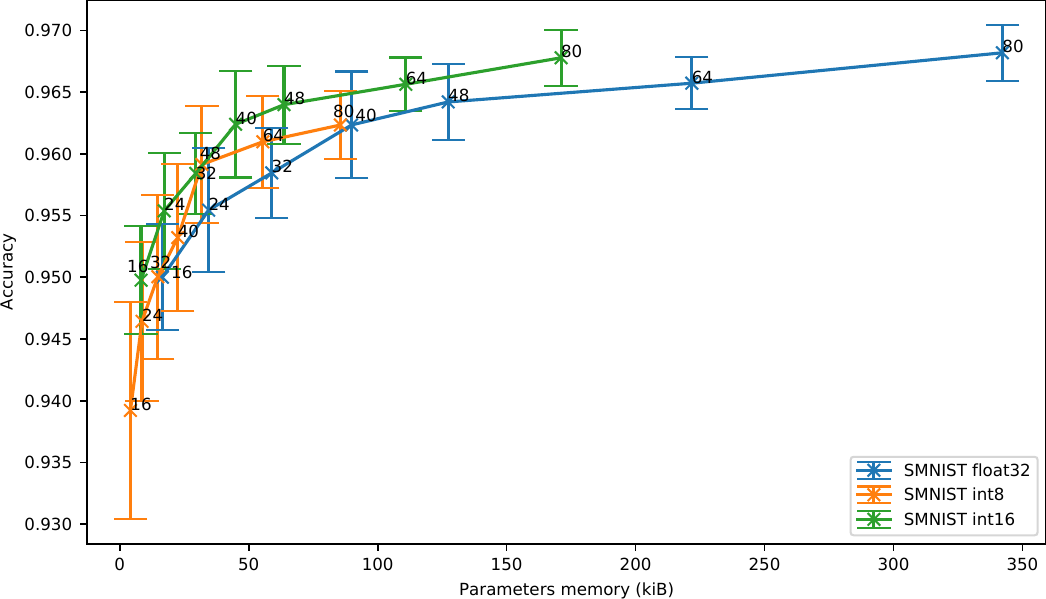}
	\caption{Spoken digits dataset (SMNIST): accuracy vs. parameter memory.} 

	\label{fig:smnistaccvsparams}
\end{figure}

In Figure~\ref{fig:smnistaccvsparams}, we can see that the 16-bit quantization is still the best solution in terms of memory footprint. Despite the fact that the 8-bit quantization stays closer to 16-bit quantization on SMNIST than on UCI-HAR, the 8-bit quantization does not provide any benefit over 16-bit quantization in terms of accuracy vs. memory ratio, even for small neural networks.

\unskip

\subsubsection{The German Traffic Sign Recognition Benchmark (GTSRB)}
The German Traffic Sign Recognition Benchmark (GTSRB~\cite{GTSRB}) is a dataset containing various color pictures of road signs.
Image sizes vary between $15\times 15$ to $250\times 250$ pixels. In this experiment, the two-dimensional images were scaled to $32\times 32$ pixels using bilinear interpolation and anti-aliasing, while keeping the 3 color channels (red, green, blue).
The dataset is divided into training and testing sets of 39,209 and 12,630 vectors, respectively.
There are 43 different classes, one for each type of road sign in the dataset.

The initial training without quantization uses a batch size of $128$ over $120$ epochs.
The initial learning rate is set to $0.01$, the momentum is set to $0.9$ and the weight decay is set to $5\times 10^{-4}$. The learning rate is multiplied by $0.1$ at epochs $40$, $80$ and $100$.

The quantization-aware training for fixed-point on 8-bit integers uses a batch size of $512$ over $120$ epochs.
The initial learning rate, momentum and weight decay are the same as for the initial training.
The learning rate is multiplied by $0.1$ at epochs $20$, $60$, $80$ and $100$.

\begin{figure}[H]
	\includegraphics[width=0.9\linewidth,page=1]{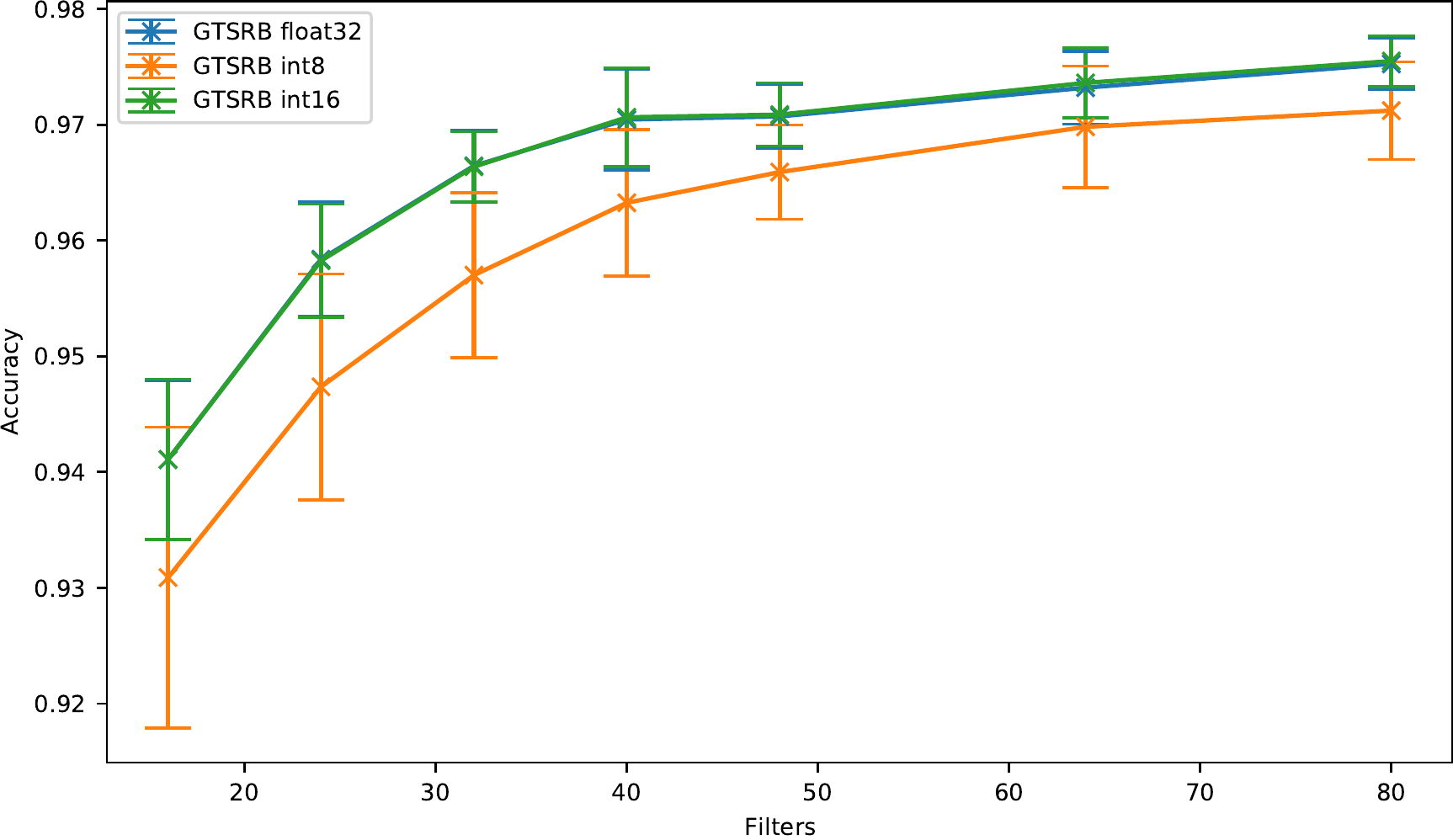}
	\caption{German Traffic Sign Recognition Benchmark: accuracy vs. filters.}

	\label{fig:gtsrbaccvsfilters}
\end{figure}

The accuracy results obtained for 8- and 16-bit quantization and the 32-bit floating-point versions are shown in Figure~\ref{fig:gtsrbaccvsfilters} for different numbers of filters. As can be seen, the \mbox{16-bit} quantization (\textit{GTSRB int16}) provides an accuracy similar to the one obtained with the baseline (\textit{GTSRB float32}). In the meantime, a drop in accuracy of up to 1.1\% can be observed when the 8-bit quantization is used with this GTSRB dataset. However, as it was observed with the SMNIST dataset, the accuracy approaches that of the baseline when the network has more filters (a drop of only 0.33\% for 64 filters). 

\begin{figure}[H]
	\includegraphics[width=0.9\linewidth,page=2]{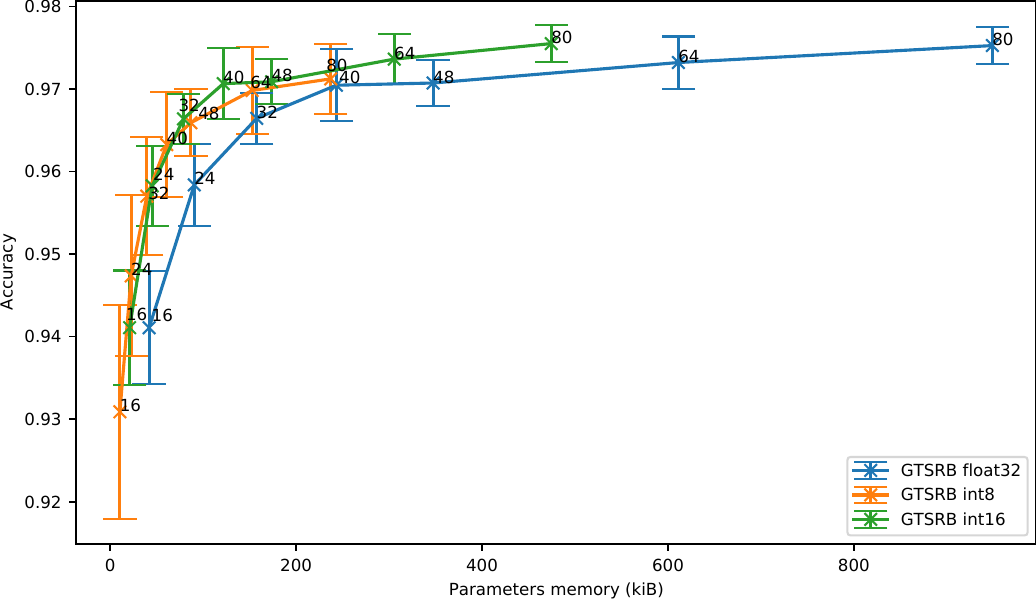}
	\caption{German Traffic Sign Recognition Benchmark: accuracy vs. parameter memory.} 

	\label{fig:gtsrbaccvsparams}
\end{figure}

Moreover, even though the 8-bit quantization does not outperform the results obtained with the 16-bit quantization, Figure~\ref{fig:gtsrbaccvsparams} shows that the 8-bit quantization can represent an interesting solution when a two-dimensional network is used on an image~dataset.

\subsection{Evaluation of Frameworks and Embedded Platforms}

In our experiments, two different targets have been used to deploy a deep neural network on a microcontroller: the SparkFun Edge and the Nucleo-L452RE-P.
Both platforms are set to run at 48~MHz on a 3.3~V supply and their main specifications are summarized in Table \ref{tab:embeddedplatforms}.

\begin{specialtable}[H]
\caption{Embedded platforms.} 
\label{tab:embeddedplatforms}
\renewcommand{\arraystretch}{1.2} 
\resizebox{\columnwidth}{!}{
\begin{tabular}{lrr}
 \noalign{\hrule height 1.0pt}
{\textbf{Board}} & {\textbf{Nucleo-L452RE-P}} & {\textbf{SparkFun Edge}} \\
\hline
MCU & STM32L452RE & Ambiq Apollo3 \\
\hline
Core & Cortex-M4F & Cortex-M4F \\
\hline
Max Clock & 80 MHz & 48 MHz (96 MHz ``Burst Mode'') \\
\hline
RAM & 128 kiB & 384 kiB \\
\hline
Flash & 512 kiB & 1024 kiB \\
\hline
CoreMark/MHz & 3.42  & 2.479 \\
\hline
Run current @3.3 V, 48 MHz & 4.80 mA & 0.82 mA * \\
 \noalign{\hrule height 1.0pt}
\end{tabular}
}
\pbox{\columnwidth}{\footnotesize~* After removing peripherals (Mic1\&2, accelerometer \ldots)}
\end{specialtable}
\vspace*{-6pt}

$V_{DD\_MCU}$ is set to 1.8~V for the Nucleo-L452RE-P platform and current measurement is taken from the $I_{DD}$ jumper. It does not have any on-board peripherals. On the SparkFun Edge board, the measure of the current is done using the power input pin of the board (after the programmer). The built-in peripherals were unsoldered from the board to eliminate their power consumption. The current consumption was measured using a Brymen BM857s auto-ranging digital multimeter configured in max mode. The energy results are based on this maximum observed current consumption and the supply voltage of 3.3~V.

As can be seen in Table \ref{tab:embeddedplatforms}, and even though both platforms are built around a Cortex-M4F core running at the same frequency, thanks to its subthreshold operation the SparkFun Edge board consumes considerably less power than the Nucleo-L452RE-P, while also having more Flash and RAM memory. However, results obtained with the CoreMark benchmark show that the Ambiq Apollo3 microcontroller is slower than the STM32L452RE. It is worth noting that the CoreMark results have been measured on the Ambiq Apollo3 microcontroller, while they have been taken from the datasheet for the STM32L452RE microcontroller.

The deep neural network used in our experiments is the residual neural network described in Section \ref{sec:results}. This network has been trained on the UCI-HAR dataset presented in Section \ref{subsubsec:ucihar}. Inference time is measured from 50 test vectors from the testing set of UCI-HAR on the microcontrollers.  TensorFlow Lite for Microcontrollers version 2.4.1 has been used to deploy the deep neural network on the SparkFun Edge board, while STM32Cube.AI version 5.2.0 has been used to deploy it on the Nucleo-L452RE-P board, both for the \mbox{32-bit} floating-point and fixed-point on 8-bit integers inference. Our framework is used to deploy the deep neural network on both platforms for 32-bit floating-point, fixed-point on 16-bit integers and fixed-point on 8-bit integer inference. It is worth noting that optimizations for the Cortex-M4F provided by CMSIS-NN are enabled for both TensorFlow Lite for Microcontrollers and STM32Cube.AI tools. Our framework does not make use of these optimizations yet. The main characteristics of the frameworks are summarized in Table \ref{tab:embeddedframeworks}. 

\begin{specialtable}[H]
\caption{Embedded AI frameworks.} 
\label{tab:embeddedframeworks}
\renewcommand{\arraystretch}{1.2} 
\resizebox{\columnwidth}{!}{
\begin{tabular}{llll}
\noalign{\hrule height 1pt}

\textbf{Framework}			& \textbf{STM32Cube.AI}					& \textbf{TFLite Micro}					& \textbf{MicroAI} \\
\hline
Source				& Keras, TFLite, \ldots		& Keras, TFLite	& Keras, PyTorch *\\
\hline
Validation			& Integrated tools				& None						& Integrated tools\\
\hline
Metrics				& RAM/ROM footprint,	            & None						    & ROM footprint\\
    				& inference time, MACC	        & 			
    				& inference time\\
\hline
Portability			& STM32 only					& Any 32-bit MCU				& Any 32-bit MCU\\
\hline
Built-in platform	& STM32 boards 			& 32F746GDiscovery, 		& SparkFun Edge, \\
support          	& (Nucleo, \ldots) 			& SparkFun Edge, \ldots		& Nucleo-L452-RE-P\\
\hline
Sources				& Private					& Public					& Public\\
\hline
Data type			& \texttt{float}, \texttt{int8\_t}					& \texttt{float}, \texttt{int8\_t}					& \texttt{float}, \texttt{int8\_t}, \texttt{int16\_t}\\
\hline
Quantized data			& Weights, activations				& Weights, activations				& Weights, activations\\
\hline
Quantizer			& Uniform (from TFlite) 			& Uniform 					& Uniform\\
\hline
Quantized coding				& Offset and scale				& Offset and scale				& Fixed-point $Qm.n$\\
\noalign{\hrule height 1pt} 
\end{tabular}}
\pbox{\columnwidth}{\footnotesize~* PyTorch models must be semi-automatically converted to Keras model prior to deployment}

\end{specialtable}

{To compare software and hardware platforms, only the results with 80 filters per convolution are analyzed below. Nevertheless, results with less than 80 filters are still available in the tables of Appendix \ref{appendix:frameworkresults} to highlight how fast and efficient a small deep neural network can be when deployed on a constrained embedded target. They also highlight a higher overhead for very small neural networks especially for TensorFlow Lite for Microcontrollers compared to our framework.}

\begin{figure}[H]
\scalebox{0.9}[0.85]{	\includegraphics[width=0.845\linewidth]{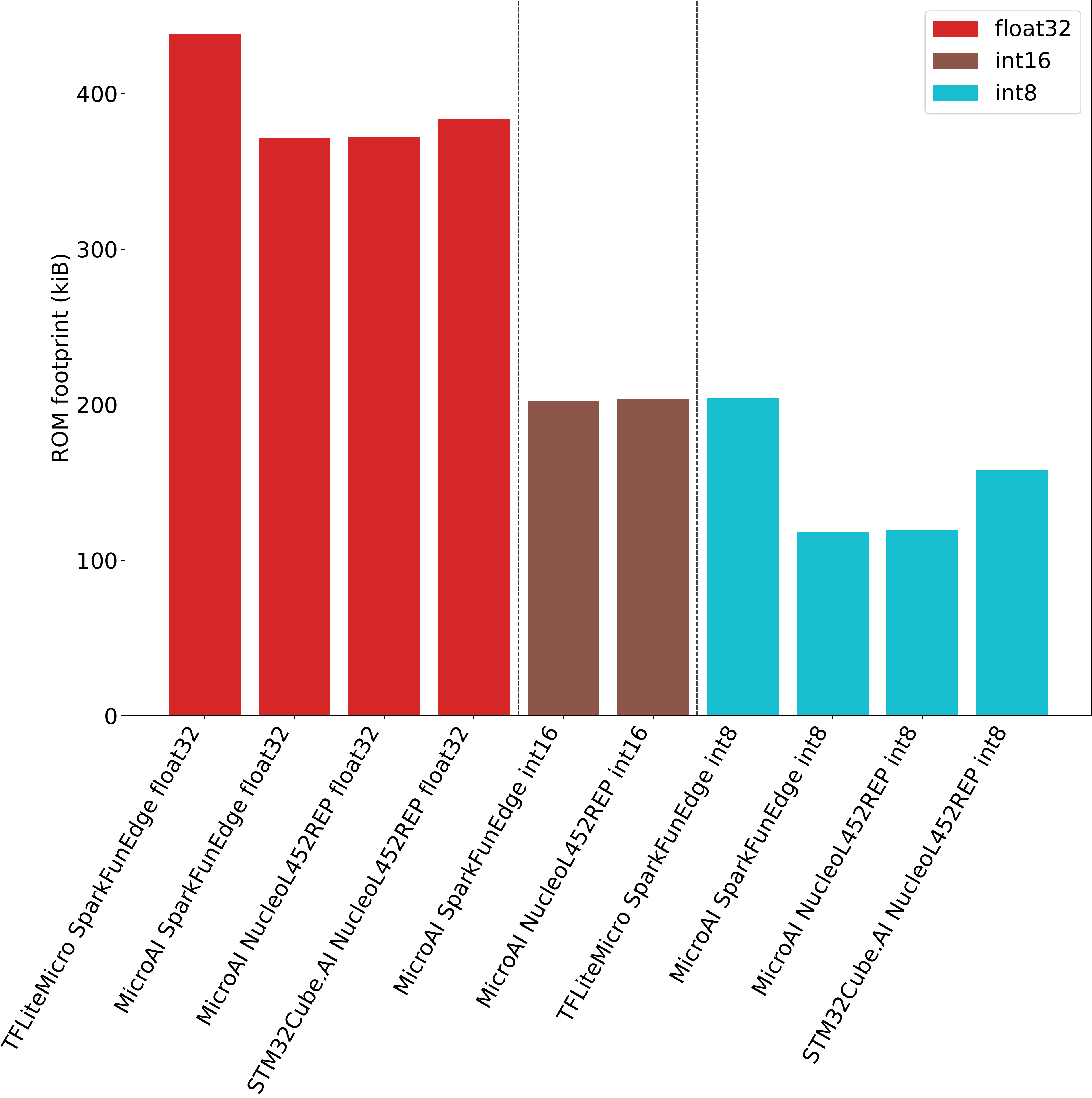}}

	\caption{\textls[-10]{ROM footprint for TFLite Micro, STM32Cube.AI and MicroAI with 80 filters per convolution.}}

	\label{fig:graphromsizevsfilters}
\end{figure}

In Figure~\ref{fig:graphromsizevsfilters}, we can observe that TFLite Micro has a higher overhead than \linebreak STM32Cube.AI, while MicroAI exhibits a slightly lower overhead than STM32Cube.AI. As outlined in Table \ref{tab:romframeworks} of Appendix \ref{appendix:frameworkresults}, when the number of filters per convolution increases, most of the ROM is used by the model's weights.

\begin{figure}[H]
\scalebox{0.9}[0.85]{	\includegraphics[width=0.845\linewidth]{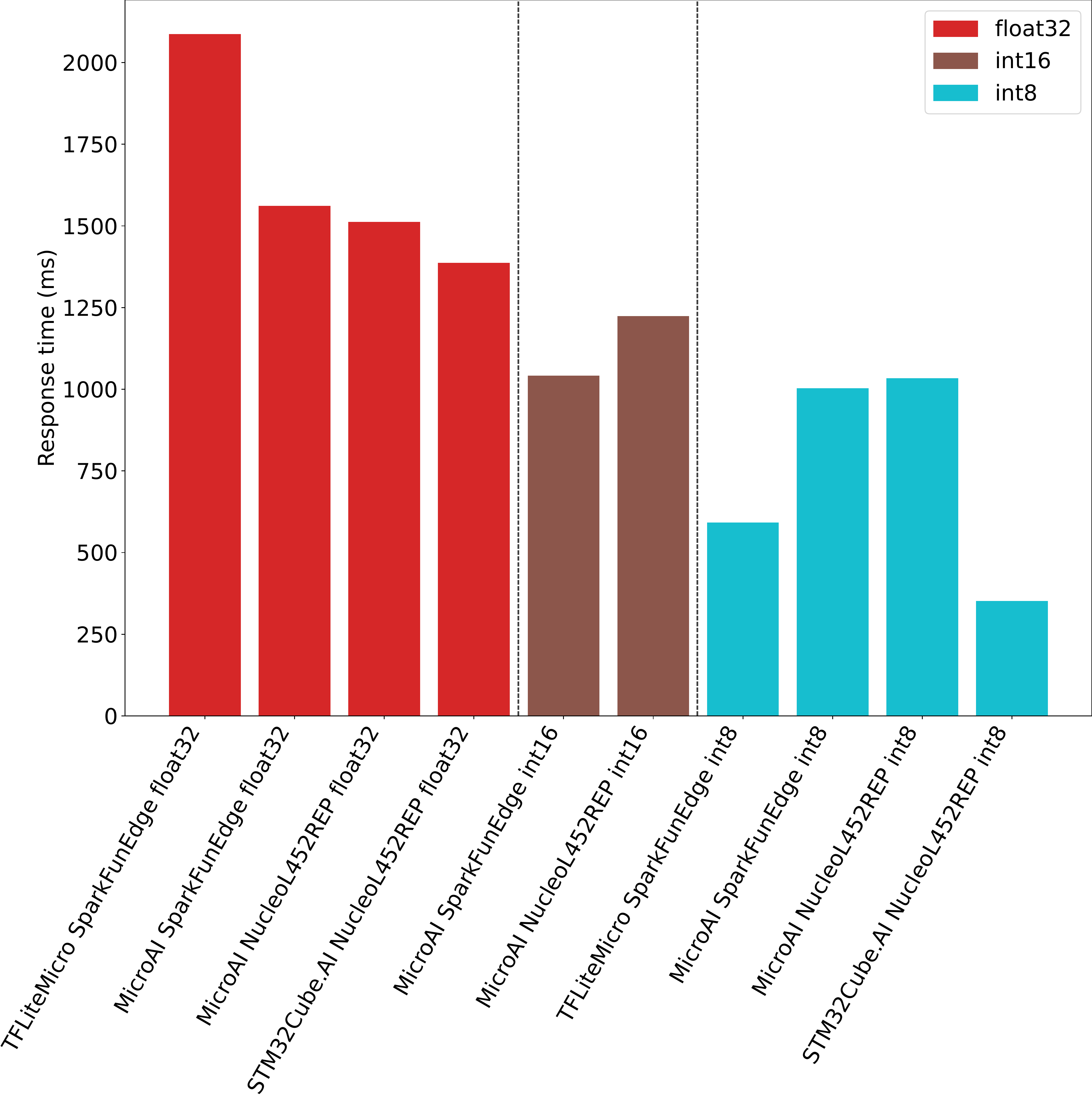}}
	\caption{Inference time for 1 input for TFLite Micro, STM32Cube.AI and MicroAI with 80 filters per convolution.}

	\label{fig:graphruntimevsfilters}
\end{figure}

The inference time obtained for both platforms and the different deployment tools is illustrated in Figure~\ref{fig:graphruntimevsfilters}.
As can be seen, the STM32Cube.AI with the 8-bit inference provides the best solution as it requires only 352~ms for one inference.
In the same configuration, TensorFlow Lite for Microcontrollers requires 592~ms for one inference. Finally, 1034~ms and 1003~ms are required for one inference using our framework on the Nucleo-L452RE-P board and the SparkFun Edge board, respectively.

\unskip

When using fixed-point on 16-bit integers for the inference, our framework provides approximately the same performance as with 8 bits. The reason is that the inference code is the same: similar instructions are generated, and computations are performed using 32-bit registers. On the Nucleo-L452RE-P, we can observe that the inference time for one input is 1223~ms, while it is only 1042~ms on the SparkFun Edge board. We guess this improvement is due to different implementations around the core in terms of memory access, especially the cache for the Flash memory.

Figure~\ref{fig:graphruntimevsfilters} also shows that, whatever the tool and target, the 32-bit floating-point inference is slower than with 16- or 8-bit quantization.
We can also observe that our framework requires 1561~ms and 1512~ms for one inference on the SparkFun Edge and the Nucleo-L452RE-P boards, respectively. The STM32Cube.AI requires \mbox{1387~ms} for one inference on the Nucleo-L452RE-P board. Our framework therefore exhibits a comparable performance to the STM32Cube.AI. 
Finally, we can see that TensorFlow Lite for microcontrollers on the SparkFun Edge board provides lower performance, requiring 2087~ms to perform one inference.

\begin{figure}[H]
	\scalebox{0.9}[0.85]{\includegraphics[width=0.845\linewidth]{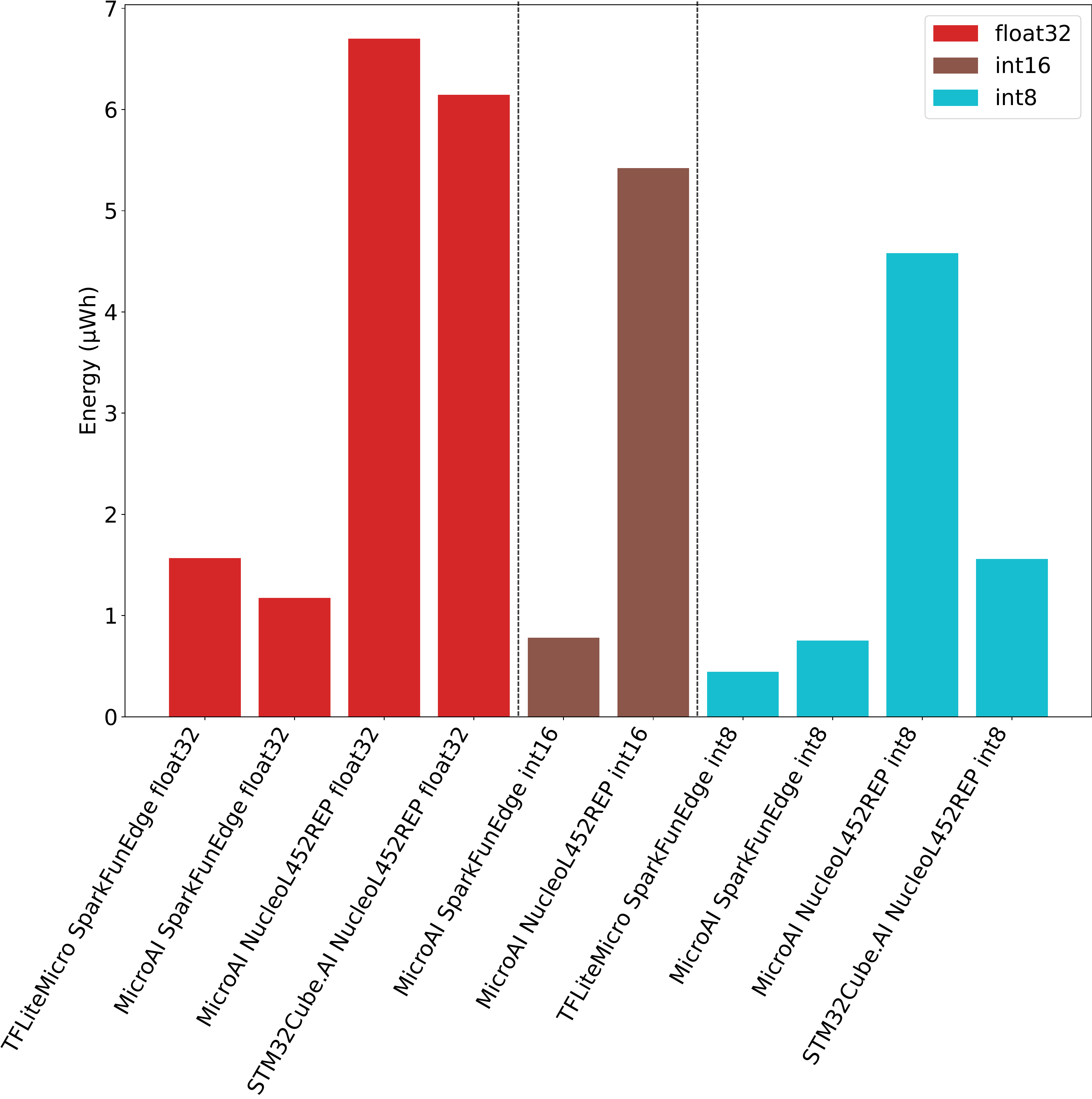}}
	\caption{Energy consumption for 1 input for TFLite Micro, STM32Cube.AI and MicroAI with 80~filters per convolution.}

	\label{fig:graphenergyvsfilters}
\end{figure}

To conclude, and as outlined in Figure~\ref{fig:graphenergyvsfilters}, we can say the SparkFun Edge board provides the best power efficiency in all situations. The reason is that the SparkFun Edge board power consumption is approximately 6 times lower than the Nucleo-L452RE-P. 
Using the SparkFun Edge board and TensorFlow Lite for Microcontroller with fixed-point on 8-bit integers, one inference requires 0.45~µWh of energy consumption. 
In contrast, our framework requires 0.75~µWh and 0.78~µWh on the SparkFun Edge board for inference with fixed-point on 8-bit and 16-bit integers, respectively.
When 32-bit floating-point is used for inference on the SparkFun Edge board, our framework provides a better energy efficiency than TensorFlow Lite for Microcontrollers as it requires 1.17~µWh instead of \mbox{1.57~µWh.}

\textls[-10]{Concerning the energy consumed on the Nucleo-L452RE-P board, our framework requires 4.58~µWh, 5.42~µWh and 6.70~µWh for one inference using fixed-point on \mbox{8-bit} integers, on 16-bit integers and 32-bit floating-point, respectively. In comparison, only \mbox{6.15~µWh} of energy is required for one inference when the STM32Cube.AI framework is used with 32-bit floating-point.
Finally, we can see that the required energy for one inference when using STM32Cube.AI with fixed-point on 8-bit integers is 1.56~µWh on the Nucleo-L452RE-P. This amount of energy is similar to the one obtained with TensorFlow Lite for Microcontrollers on the SparkFun Edge board when performing floating-point~inference.} 

\section{Discussion}
\label{sec:discussion}

First, a high variance is observable when we compare the accuracy results obtained on the three datasets versus the model size. This variability makes it difficult to draw any definitive conclusions.
However, there is a trend in our results that provides some insights into performance for each experiment.

As has been shown, execution using fixed-point on 8-bit and 16-bit integers provides a significant decrease in the inference time, thus also reducing the average power consumption. As power consumption is a key parameter in embedded systems, shorter inference times are interesting as they make it possible either to reduce the microcontroller's operating frequency or to put the microcontroller in sleep mode for a longer period between two inferences. 
In addition, execution using 8-bit and 16-bit integers also provides a significant reduction in memory footprint. The memory required for the model parameters is divided by 4 and 2 for for 8-bit and 16-bit quantization, respectively. It is worth noting that the RAM usage, which is not illustrated here, is also reduced.

Our results also show that performing inference using quantization with fixed-point on 16-bit integers does not lead to a drop in accuracy, whatever test case is considered.  Moreover, inference using 16 bits does not require quantization-aware training to achieve such results. 
As both the power consumption and the memory footprint can be decreased, fixed-point quantization on 16-bit integers is therefore always preferable to 32-bit floating-point inference.

Conversely, 8-bit quantization does not provide a substantial improvement over 16-bit quantization. Moreover, 8-bit quantization requires performing quantization-aware training.
It is worth noting that quantization-aware training for 8-bit quantization introduces more variance in the results over the baseline, and is also more sensitive to a change in the training parameters.
As it is quite difficult to achieve a stable training, it is preferable to use an optimizer such as SGD with conservative parameters, instead of optimizers such as Adam or RAdam, to reduce the variance of the results, even though it  means achieving a lower maximum accuracy.

During our experiments, it was also observed that the 8-bit post-training quantization of TensorFlow Lite achieved better results compared to the 8-bit quantization-aware training provided by our framework. This is likely due to the combination of per-filter quantization, asymmetric range and non-power-of-two scale factor, as well as optimizations of TensorFlow Lite to avoid unnecessary truncation and thus loss of precision. 
We also observed that using 9 bits instead of 8 bits during the post-training quantization allows us to outperform the TensorFlow Lite quantization performance. Some results showing this improvement are available in Appendix \ref{appendix:tfliteuquantmicroai9} for the UCI-HAR dataset. 
From these results, we can conclude that the slight additional precision brought by the combination of per-filter quantization, asymmetric range and non-power of two scale factor does in fact matter. 
Implementing these methods in our framework seems therefore required to reduce the accuracy loss of our 8-bit quantization.

Another benefit of 8-bit quantization is that SIMD instructions can be used (with some classes of microcontrollers) to improve the inference time and thus further reduce the power consumption.
Such instructions allow performing in a single cycle either 2 multiply–accumulate operations with \mbox{16-bit} operands and a common accumulator (\textit{SMLAD}), or 2 additions of \mbox{16-bit} operands (\textit{QADD16}), or a shift and saturation operation (\textit{SSAT}).
The \textit{SMLAD}, \textit{QADD16} and \textit{SSAT} instructions are not yet used in our framework, but this work is in progress.
Nonetheless, a 16-bit quantization scheme can be used with our framework, which is not the case with either TensorFlow Lite for Microcontrollers or STM32Cube.AI. As presented in the results, the 16-bit quantization from our framework provides a good compromise between accuracy, inference time and memory footprint, without requiring additional work on quantization-aware training.

The results obtained on inference time clearly show that both the software and hardware platforms have a substantial impact on energy efficiency.
STM32Cube.AI offers the most optimized inference engine in terms of execution time, both in floating-point and fixed-point on integers. Our results show that TensorFlow Lite for Microcontrollers is slower than STM32Cube.AI in both conditions.
For the floating-point inference, our framework is in between these two software platforms and is only slightly slower than STM32Cube.AI. However, as optimizations using SIMD instructions have not been implemented yet in our framework, inference using 8-bit integers still provides lower performance than TensorFlow Lite for Microcontrollers and STM32Cube.AI.

\textls[-5]{Regardless of the software performance, running STM32Cube.AI on a Nucleo-L452RE-P board is only competitive with inference using 8-bit integers when compared to TensorFlow Lite for Microcontrollers in 32-bit floating-point inference on the SparkFun Edge board. The reason is that the Ambiq Apollo3 microcontroller on the SparkFun Edge board is much more energy efficient.
In all remaining cases, running TensorFlow Lite for Microcontrollers or our framework on the SparkFun Edge board provides much better energy efficiency figures than running STM32Cube.AI or our framework on the Nucleo-L452RE-P~board.}

\section{Conclusions}
\label{sec:conclusion}

In this work, we presented a framework to perform quantization and then deployment of deep neural networks on microcontrollers. This framework represents an alternative to the STM32Cube.AI proprietary solution and TensorFlow Lite for Microcontrollers, an open-source but complex environment. 
Inference time and energy efficiency measured on two different embedded platforms demonstrated that our framework is a viable alternative to the aforementioned solutions to perform deep neural network inference.
Our framework also introduces a fixed-point on 16-bit integer post-training quantization which is not available with the two other frameworks. We have shown that this 16-bit fixed-point quantization provides an improvement over a 32-bit floating-point inference, while being competitive with fixed-point on 8-bit integer quantization-aware training.
It provides a reduced inference time compared to floating-point inference. Moreover, the memory footprint is divided by two while keeping the same accuracy. The 8-bit quantization provides further improvements in inference time and memory footprint but at the cost of a slight decrease in accuracy and a more complex implementation.

Work is still in progress to implement some optimization techniques for fixed-point on 8-bit integer inference. Three optimizations are especially targeted: per-filter quantization, asymmetric range and non-power-of-two scale factor. In addition, using SIMD instructions in the inference engine should help further decrease the inference time. These optimizations would therefore make our framework more competitive in terms of inference time and accuracy. 
Another possible improvement for fixed-point on integers inference consists of using 8-bit quantization for the weights and 16-bit quantization for the activations. TensorFlow Lite for Microcontrollers is currently in the process of implementing this technique. 
Mixed precision can indeed provide a way to reduce the memory footprint of layers that do not need a high-precision representation (using 8 bits for weights and activations), while keeping a higher precision (16-bit representation) for layers that need it. The CMix-NN~\cite{CmixNN} library already provides an implementation of convolution functions for various data type configurations (in 2, 4 and 8 bits).
To further improve power consumption and memory footprint, binary neural networks can also be considered. However, to run them efficiently on microcontrollers, binary neural networks would need to be implemented using bit-wise operations on 32-bit registers. This way, as many as 32 computations could be performed in parallel.

Apart from quantization, other techniques can be used to improve the execution of deep neural networks on embedded targets. One of these techniques is the big/LITTLE DNN approach~\cite{bigLITTLE} where the inference is first done on a very small deep neural network. Then, if the confidence is too low, inference is done using a larger deep neural network to reduce the confusion of the classification task.
This technique allows a fast inference response time for most inputs, thus lowering the power consumption. In fact, it has been shown that the set of inputs that are difficult to classify and so require running the bigger deep neural network is small.
However, this approach does not lower the memory footprint. Other techniques such as pruning can also be used to obtain a smaller deep neural network while keeping the same accuracy.
When structured pruning~\cite{StructPruning} is used, for instance, entire filters are completely removed from the convolutional neural network model. This reduces both the memory footprint and the power consumption. 
Finally, other optimization techniques also consider new neural network architectures. One can cite for example the recently published MCUNet~\cite{mcunet} framework with its TinyNAS tool that aims to identify the neural network model that will best perform on the target.

Future works will also be dedicated to the deployment of neural network architectures on FPGA using high-level synthesis tools such as Vivado. In fact, a feasibility study has already been performed and has shown that our framework can be also used for deployment on FPGA. 
Moreover, work is in progress to natively support automatic PyTorch deployment. To do so, the features provided by the \texttt{torch.fx} module of the newly released PyTorch 1.8.0 are used.

Finally, we are currently working on a real application of our framework that consists in integrating artificial intelligence into smart glasses \cite{smartglasses} to perform, among other tasks, human activity recognition in the context of elder care. Preliminary results have been published in \cite{DSD2020}.

\vspace{6pt} 


\supplementary{The open-source MicroAI software framework~\cite{microaisoftware} is available online at \url{https://bitbucket.org/edge-team-leat/microai\_public}.}

\authorcontributions{Investigation, P.N.; methodology, P.N. and G.B.H.; software, P.N. and G.B.H.; supervision, A.P., B.M. and V.G.; writing---original draft preparation, P.N.; writing---review and editing, G.B.H., A.P., B.M., V.G. All authors have read and agreed to the published version of the manuscript.}

\funding{This research is funded by ``Universit\'e C\^ote d'Azur'', ``CNRS'', ``R\'egion Sud Provence-Alpes-C\^ote d'Azur, France''  {and ``Ellcie Healthy''}.}

\institutionalreview{Not applicable.}

\informedconsent{Not applicable.}



\conflictsofinterest{The authors declare no conflict of interest. The funders had no role in the design of the study; in the collection, analyses, or interpretation of data; in the writing of the manuscript, or in the decision to publish the~results.}



\appendixtitles{yes} 
\appendixstart
\appendix

\section{Comparison of the Inference Times of a Microcontroller, a CPU and a GPU}
\label{appendix:mcucpugpubench}

\begin{specialtable}[H]
\caption{Microcontroller (STM32L452RE), CPU (Intel Core i7-8850H) and GPU (NVidia Quadro P2000M) platforms. Power consumption figures for the GPU and the CPU are the TDP values from the manufacturer and do not reflect the exact power consumption of the device.} 
\renewcommand{\arraystretch}{1.2}
\resizebox{\columnwidth}{!}{
\begin{tabular}{cccc}
\noalign{\hrule height 1pt} 
\textbf{Platform} & \textbf{Model}	& \textbf{Framework}	& \textbf{Power Consumption}	\\

\hline
MCU & STM32L452RE		& STM32Cube.AI	& 0.016 W	\\
\hline
CPU & Intel Core i7-8850H	& TensorFlow	& 45 W			\\
\hline
GPU & NVidia Quadro P2000M	& TensorFlow	& 50 W			\\
\noalign{\hrule height 1pt} 
\end{tabular}}
\end{specialtable}
\unskip

\begin{specialtable}[H]
\caption{Comparison of 32-bit floating-point inference time for a single input on a microcontroller, a CPU and a GPU. The neural network architecture is described in Section \ref{sec:results} with the number of filters per convolution layer varying from 16 to 80, and the dataset is described in Section \ref{subsubsec:ucihar}. For the CPU and the GPU, the inference batch size is set to 512 and the dataset is repeated 104 times to try to compensate for the large startup overhead compared to the total inference time. Measurements are averaged over at least 5 runs.}
\renewcommand{\arraystretch}{1.2} 
\resizebox{\columnwidth}{!}{
\begin{tabular}{cccccccc}
\noalign{\hrule height 1pt}
 & \multicolumn{7}{c}{\textbf{Inference Time (ms)}}\\
\cline{2-8}
\multirow{-2}{*}{\textbf{Platform}} & \textbf{16 Filters} & \textbf{24 Filters} & \textbf{32 Filters} & \textbf{40 Filters} & \textbf{48 Filters} & \textbf{64 Filters} & \textbf{80 Filters}\\
\hline 
MCU		& $85$ & $174$ & $271$ & $404$ & $544$ & $921$ & $1387$\\
\hline 
CPU	& $0.0396$ & $0.0552$ & $0.0720$ & $0.0937$ & $0.1134$ & $0.1538$ & $0.2046$\\
\hline 
GPU	& $0.0227$ & $0.0197$ & $0.0223$ & $0.0284$ & $0.0317$ & $0.0395$ & $0.0515$\\
\noalign{\hrule height 1pt} 
\end{tabular}}
\end{specialtable}

\section{{Number of Integer ALU Operations for a Fixed-Point Residual Neural~Network}}
\label{appendix:aluintops}

\vspace{-6pt}

\begin{specialtable}[H]
\caption{Number of arithmetic and logic operations with fixed-point on integers inference for the main layers of a residual neural network with $f$ the number of filters (output channels), $s$ the number of input samples, $c$ the number of input channels, $k$ the kernel size, $n$ the number of neurons and $i$ the number of input layers to the residual Add layer. Conv1D is assumed to be without padding and with a stride of 1.}
\label{tab:aluintops}
\renewcommand{\arraystretch}{1.2} 
\resizebox{\columnwidth}{!}{
\begin{tabular}{lllll}
\noalign{\hrule height 1pt}
 & \textbf{MACC (1 Cycle)} & \textbf{Add (1 Cycle)}         & \textbf{Shift (1 Cycle)}       & \textbf{Max/Saturate (2 Cycles)}\\
\hline  
Conv1D          & $f \times s \times c \times k$ & N/A                   & $2\times f \times s$                   & $f \times s$\\
\hline
ReLU         & N/A           & N/A                   & N/A                 & $c \times s$\\
\hline
Maxpool         & N/A           & N/A                   & N/A                 & $c \times s \times k$\\
\hline
Add             & N/A           & $s \times c \times (i - 1)$       & $s \times c \times i$             & $c \times s$\\
\hline
FullyConnected  & $n \times s$         & N/A                   & $2 \times n$                     & $n$\\
\noalign{\hrule height 1pt} 
\end{tabular}
}
\end{specialtable}

\section{Comparison of TensorFlow Lite for Microcontrollers and MicroAI~Quantizations}
\label{appendix:tfliteuquantmicroai9}

\begin{figure}[H]
	\includegraphics[width=0.95\linewidth,page=1]{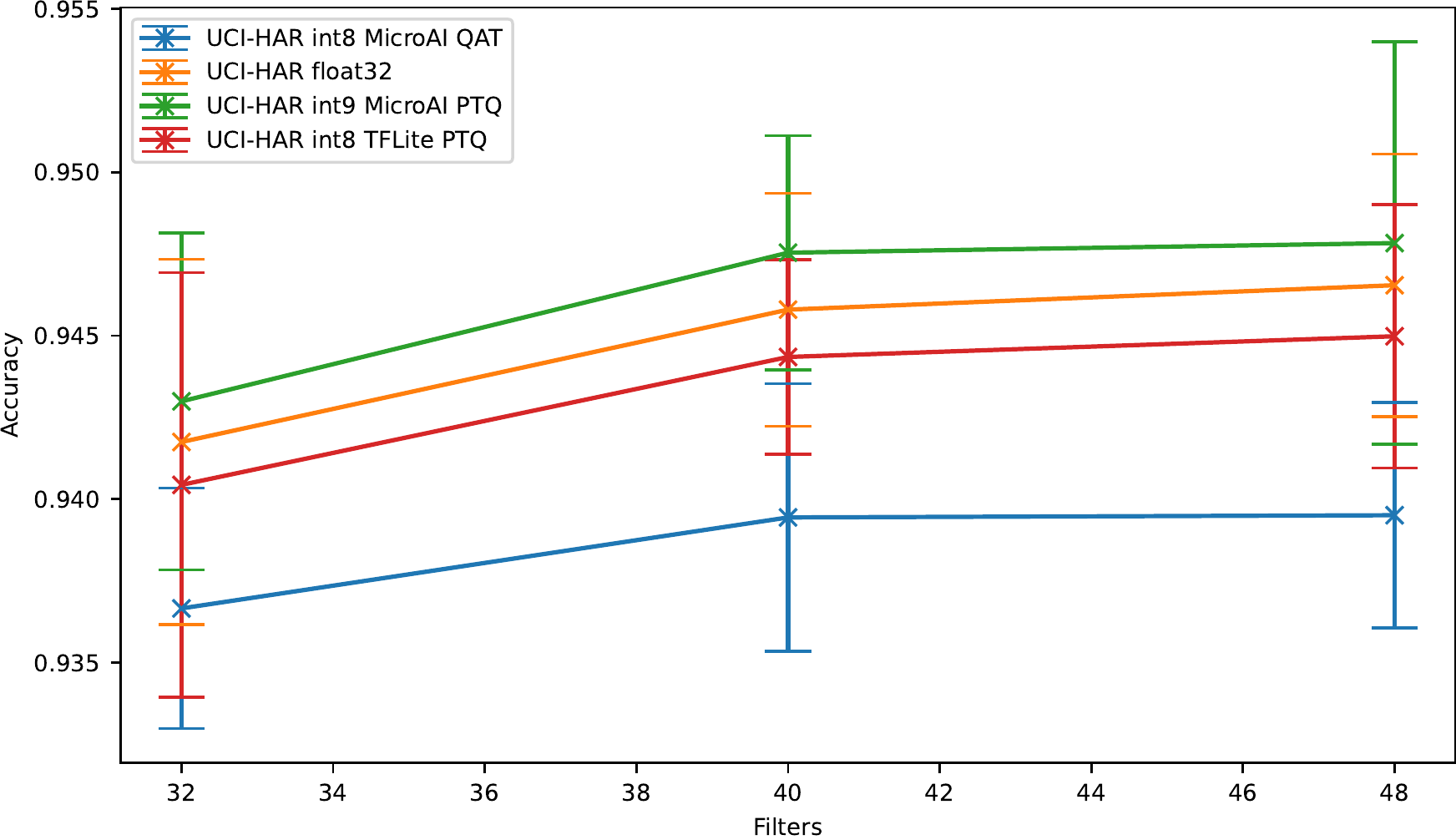}
	\caption{Accuracy vs. filters for baseline (\textit{float32}), 8-bit post-training quantization from TensorFlow Lite (\textit{int8 TFLite PTQ}), 8-bit quantization-aware training from our framework (\textit{int8 MicroAI QAT}), and 9-bit post-training quantization from our framework (\textit{int9 MicroAI PTQ}). The neural network architecture is described in Section \ref{sec:results} with the number of filters per convolution layer varying from 32 to 48, and the dataset is described in Section \ref{subsubsec:ucihar}.}

\end{figure}

\section{MicroAI Commands to Run for Automatic Training and Deployment of Deep Neural Networks}
\label{appendix:microaicommands}

Data can be preprocessed (e.g., to apply normalization) from the source dataset and serialized to an intermediate dataset file with the following command:

\begin{lstlisting}[language=Bash]
microai <config.toml> preprocess_data
\end{lstlisting}
\vspace*{-12pt}

The training phase is started by running the following command:

\begin{lstlisting}[language=Bash]
microai <config.toml> train
\end{lstlisting}
\vspace*{-12pt}

Before being deployed and evaluated, the appropriate code must be generated and built for the targeted platform by running the following command:

\begin{lstlisting}[language=Bash]
microai <config.toml> prepare_deploy
\end{lstlisting}
\vspace*{-12pt}

Once the binaries are generated, they can be deployed, and the model can be evaluated on the target by running the following command:

\begin{lstlisting}[language=Bash]
microai <config.toml> deploy_and_evaluate
\end{lstlisting}
\vspace*{-24pt}

\pagebreak
\section{ {Detailed Results of the Evaluation of Frameworks and Embedded Platforms Evaluation}}
\label{appendix:frameworkresults}

\end{paracol}
\nointerlineskip
\appendix
\begin{specialtable}[H]
\widetable
\caption{ROM footprint vs. filters for TFLite Micro, STM32Cube.AI and MicroAI.}
\label{tab:romframeworks}
\renewcommand{\arraystretch}{1.2} 
\small
\begin{tabular}{lllrrrrrrr}
\noalign{\hrule height 1pt} 
\multicolumn{3}{c}{} &\multicolumn{7}{c}{\textbf{ROM Footprint (kiB)}}\\
\hline
\textbf{Framework} & \textbf{Target} & \textbf{Data Type} & \textbf{16 Filters} & \textbf{24 Filters} & \textbf{32 Filters} & \textbf{40 Filters} & \textbf{48 Filters} & \textbf{64 Filters} & \textbf{80 Filters}\\
\hline
TFLiteMicro & SparkFunEdge & float32 & 116.520 & 133.988 & 157.957 & 188.426 & 225.395 & 318.926 & 438.363\\
\hline
MicroAI & SparkFunEdge & float32 & 54.316 & 67.066 & 91.035 & 121.512 & 158.473 & 251.863 & 371.332\\
\hline
MicroAI & NucleoL452REP & float32 & 55.770 & 68.145 & 92.129 & 122.582 & 159.559 & 253.004 & 372.434\\
\hline
STM32Cube.AI & NucleoL452REP & float32 & 61.965 & 79.449 & 103.410 & 133.898 & 170.859 & 264.289 & 383.742\\
\hline
MicroAI & SparkFunEdge & int16 & 46.952 & 50.629 & 62.629 & 77.832 & 96.355 & 142.973 & 202.699\\
\hline
MicroAI & NucleoL452REP & int16 & 48.129 & 51.629 & 63.613 & 78.855 & 97.340 & 144.051 & 203.770\\
\hline
TFLiteMicro & SparkFunEdge & int8 & 111.051 & 117.066 & 124.691 & 133.957 & 144.832 & 171.473 & 204.613\\
\hline
MicroAI & SparkFunEdge & int8 & 43.256 & 42.249 & 48.229 & 55.854 & 65.089 & 88.343 & 118.202\\
\hline
MicroAI & NucleoL452REP & int8 & 45.038 & 43.474 & 49.464 & 57.078 & 66.322 & 89.683 & 119.541\\
\hline
STM32Cube.AI & NucleoL452REP & int8 & 72.742 & 77.746 & 84.336 & 92.582 & 102.430 & 126.996 & 158.098\\
\noalign{\hrule height 1pt} 
\end{tabular}

\end{specialtable}
\unskip

\begin{specialtable}[H]
\widetable
\caption{Inference time for one input vs. filters for TFLite Micro, STM32Cube.AI and MicroAI.}
\label{tab:timeframeworks}
\small
\renewcommand{\arraystretch}{1.2} 
\begin{tabular}{lllrrrrrrr}
\noalign{\hrule height 1pt} 
\multicolumn{3}{c}{} &\multicolumn{7}{c}{\textbf{Response Time (ms)}}\\
\hline 
\textbf{Framework} & \textbf{Target} & \textbf{Data Type} & \textbf{16 Filters} & \textbf{24 Filters} & \textbf{32 Filters} & \textbf{40 Filters} & \textbf{48 Filters} & \textbf{64 Filters} & \textbf{80 Filters}\\
\hline 
TFLiteMicro & SparkFunEdge & float32 & 179.633 & 294.157 & 438.541 & 624.172 & 860.835 & 1406.945 & 2087.241\\
\hline 
MicroAI & SparkFunEdge & float32 & 53.247 & 153.732 & 259.212 & 394.494 & 569.852 & 1017.118 & 1561.264\\
\hline 
MicroAI & NucleoL452REP & float32 & 55.762 & 152.426 & 259.160 & 395.721 & 559.249 & 976.732 & 1512.143\\
\hline 
STM32Cube.AI & NucleoL452REP & float32 & 85.359 & 174.082 & 271.362 & 403.898 & 544.406 & 921.646 & 1387.083\\
\hline 
MicroAI & SparkFunEdge & int16 & 40.867 & 113.035 & 191.439 & 287.655 & 389.450 & 667.547 & 1041.617\\
\hline 
MicroAI & NucleoL452REP & int16 & 44.915 & 120.308 & 205.499 & 318.310 & 459.880 & 796.310 & 1223.513\\
\hline 
TFLiteMicro & SparkFunEdge & int8 & 92.529 & 130.760 & 172.673 & 225.092 & 280.942 & 418.198 & 591.785\\
\hline 
MicroAI & SparkFunEdge & int8 & 39.417 & 101.704 & 172.551 & 259.830 & 375.840 & 658.441 & 1003.365\\
\hline 
MicroAI & NucleoL452REP & int8 & 43.003 & 107.705 & 180.830 & 272.986 & 383.761 & 659.996 & 1034.033\\
\hline 
STM32Cube.AI & NucleoL452REP & int8 & 32.297 & 53.871 & 80.388 & 111.635 & 146.022 & 242.002 & 352.079\\
\noalign{\hrule height 1pt} 
\end{tabular}

\end{specialtable}
\unskip

\begin{specialtable}[H]
\widetable
\caption{Energy consumption for 1 input vs. filters for TFLite Micro, STM32Cube.AI and MicroAI.}
\label{tab:energyframeworks}
\small
\renewcommand{\arraystretch}{1.2} 
\begin{tabular}{lllrrrrrrr}
\noalign{\hrule height 1pt} 
\multicolumn{3}{c}{} &\multicolumn{7}{c}{\textbf{Energy (µWh)}}\\
\hline
\textbf{Framework} & \textbf{Target} & \textbf{Data Type}& \textbf{16 Filters} & \textbf{24 Filters} & \textbf{32 Filters} & \textbf{40 Filters} & \textbf{48 Filters} & \textbf{64 Filters} & \textbf{80 Filters}\\
\hline 
TFLiteMicro & SparkFunEdge & float32 & 0.135 & 0.221 & 0.330 & 0.469 & 0.647 & 1.058 & 1.569\\
\hline 
MicroAI & SparkFunEdge & float32 & 0.040 & 0.116 & 0.195 & 0.297 & 0.428 & 0.765 & 1.174\\
\hline 
MicroAI & NucleoL452REP & float32 & 0.247 & 0.675 & 1.148 & 1.753 & 2.478 & 4.327 & 6.700\\
\hline 
STM32Cube.AI & NucleoL452REP & float32 & 0.378 & 0.771 & 1.202 & 1.789 & 2.412 & 4.083 & 6.146\\
\hline 
MicroAI & SparkFunEdge & int16 & 0.031 & 0.085 & 0.144 & 0.216 & 0.293 & 0.502 & 0.783\\
\hline 
MicroAI & NucleoL452REP & int16 & 0.199 & 0.533 & 0.910 & 1.410 & 2.038 & 3.528 & 5.421\\
\hline 
TFLiteMicro & SparkFunEdge & int8 & 0.070 & 0.098 & 0.130 & 0.169 & 0.211 & 0.314 & 0.445\\
\hline 
MicroAI & SparkFunEdge & int8 & 0.030 & 0.076 & 0.130 & 0.195 & 0.283 & 0.495 & 0.754\\
\hline 
MicroAI & NucleoL452REP & int8 & 0.191 & 0.477 & 0.801 & 1.209 & 1.700 & 2.924 & 4.581\\
\hline 
STM32Cube.AI & NucleoL452REP & int8 & 0.143 & 0.239 & 0.356 & 0.495 & 0.647 & 1.072 & 1.560\\
\noalign{\hrule height 1pt} 
\end{tabular}

\end{specialtable}


\reftitle{References}


\end{document}